\documentclass{article}

\usepackage[preprint]{cpal_2025}

\usepackage{url}
\usepackage{graphicx}
\usepackage{amssymb}
\usepackage{amsmath}

\usepackage{booktabs}
\usepackage{multirow}

\usepackage{hyperref}
\hypersetup{
    colorlinks=true,
    linkcolor=blue,
    filecolor=magenta,      
    urlcolor=cyan,
    }

\newtheorem{theorem}{Theorem}
\newtheorem{assumption}{Assumption}
\newtheorem{definition}{Definition}
\newtheorem{lemma}{Lemma}

\title{Latent Manifold Reconstruction and Representation with Topological and Geometrical Regularization}

\author{%
  Ren Wang\textsuperscript{1,2}, ~Pengcheng Zhou\textsuperscript{2, 3} \\
  \textsuperscript{1}Southern University of Science and Technology\\
  \textsuperscript{2}Shenzhen University of Advanced Technology\\
  \textsuperscript{3}Shenzhen Institute of Advanced Technology, Chinese Academy of Sciences\\
  \texttt{r.wang3@siat.ac.cn}
}

\begin{document}

\maketitle

\begin{abstract}
Manifold learning aims to discover and represent low-dimensional structures underlying high-dimensional data while preserving critical topological and geometric properties.
Existing methods often fail to capture local details with global topological integrity from noisy data or construct a balanced dimensionality reduction, resulting in distorted or fractured embeddings.
We present an AutoEncoder-based method that integrates a manifold reconstruction layer, which uncovers latent manifold structures from noisy point clouds,
and further provides regularizations on topological and geometric properties during dimensionality reduction,
whereas the two components promote each other during training.
Experiments on point cloud datasets demonstrate that our method outperforms baselines like t-SNE, UMAP, and Topological AutoEncoders in discovering manifold structures from noisy data and preserving them through dimensionality reduction, as validated by visualization and quantitative metrics.
This work demonstrates the significance of combining manifold reconstruction with manifold learning to achieve reliable representation of the latent manifold,
particularly when dealing with noisy real-world data.
Code repository: \url{https://github.com/Thanatorika/mrtg}.
\end{abstract}

\section{Introduction}

\textbf{Background.}
In the area of manifold learning,
or nonlinear dimensionality reduction \cite{ManifoldLearning},
one foundational assumption is the so-called "manifold hypothesis":
high dimensional data tend to locate near a low-dimensional manifold \cite{FMN16}.
In real life,
the ground truth of the low-dimensional data manifold is often implicit or \emph{latent}.
One major task of manifold learning is to explicitly transform or \textit{embed} high-dimensional noisy data into low-dimensional spaces,
while keeping the topological and geometric properties of the underlying data manifold.

Several manifold learning methods, including Isomap \cite{Isomap}, t-SNE \cite{t-SNE}, and UMAP \cite{UMAP},
use the neighborhood of a data point to approximate its tangent space,
thus keeping some local structures of neighboring data points.
However, these methods do not explicitly regulate the global topology of the latent manifold, and their embeddings are often fractured as clusters.
To address this, \citet{TopoAE} introduced Topological AutoEncoders, which preserve the global shape by incorporating a topological regularizer, 
although local structures still experience distortion or stretching. 
Also, because of the noise in the raw data, directly applying manifold structural constraints can mislead the regularization terms in the dimensionality reduction process.

\textbf{Approach.}
In this paper, we propose integrating a manifold reconstruction process with an AutoEncoder-based manifold learning approach to achieve improved manifold reconstruction and representation. The manifold reconstruction step is designed to capture the intrinsic global and local structures of the data, which can then be used as regularization constraints for the AutoEncoder in obtaining low-dimensional representations. As the training is conducted in an end-to-end manner, the hyperparameters for the manifold reconstruction process will also be optimized, resulting in a mutual-promotion between the manifold reconstruction and representation components.

\textbf{Implementation.}
In our paper,
we present a simple implementation of this idea with an AutoEncoder \cite{DeepLearning} equipped with a manifold reconstruction layer,
a topological regularizer,
and a geometrical regularizer.
We use the manifold reconstruction layer to transform data points to the latent manifold, by contracting a point to the expected direction and distance of the closest point on the latent manifold, based on the "manifold fitting" algorithm proposed by \citet{YSLY23}.
We use the AutoEncoder network to learn a low-dimensional latent representation, and utilize the encode-decode loss with topological and geometrical regularizers to optimize the representation quality.
For the topological regularizer, we use topological signature loss \cite{TopoAE} based on persistent homology \cite{PH}.
For the geometrical regularizer, we use a relaxed distortion measure \cite{RelaxedDistortionMeasure} rooted in the (scaled) isometry between Riemannian manifolds..

\textbf{Validation.}
We experiment with our proposed approach and implement the method on 3D and high-dimensional point cloud datasets,
and validate them with visualization and evaluation metrics measuring the local and global similarity between the original point cloud and the low-dimensional representation.
We also perform ablation study to validate our model components and their mutual-promotion virtue.

\begin{figure}[t]
    \centering
    \includegraphics[width=1\linewidth]{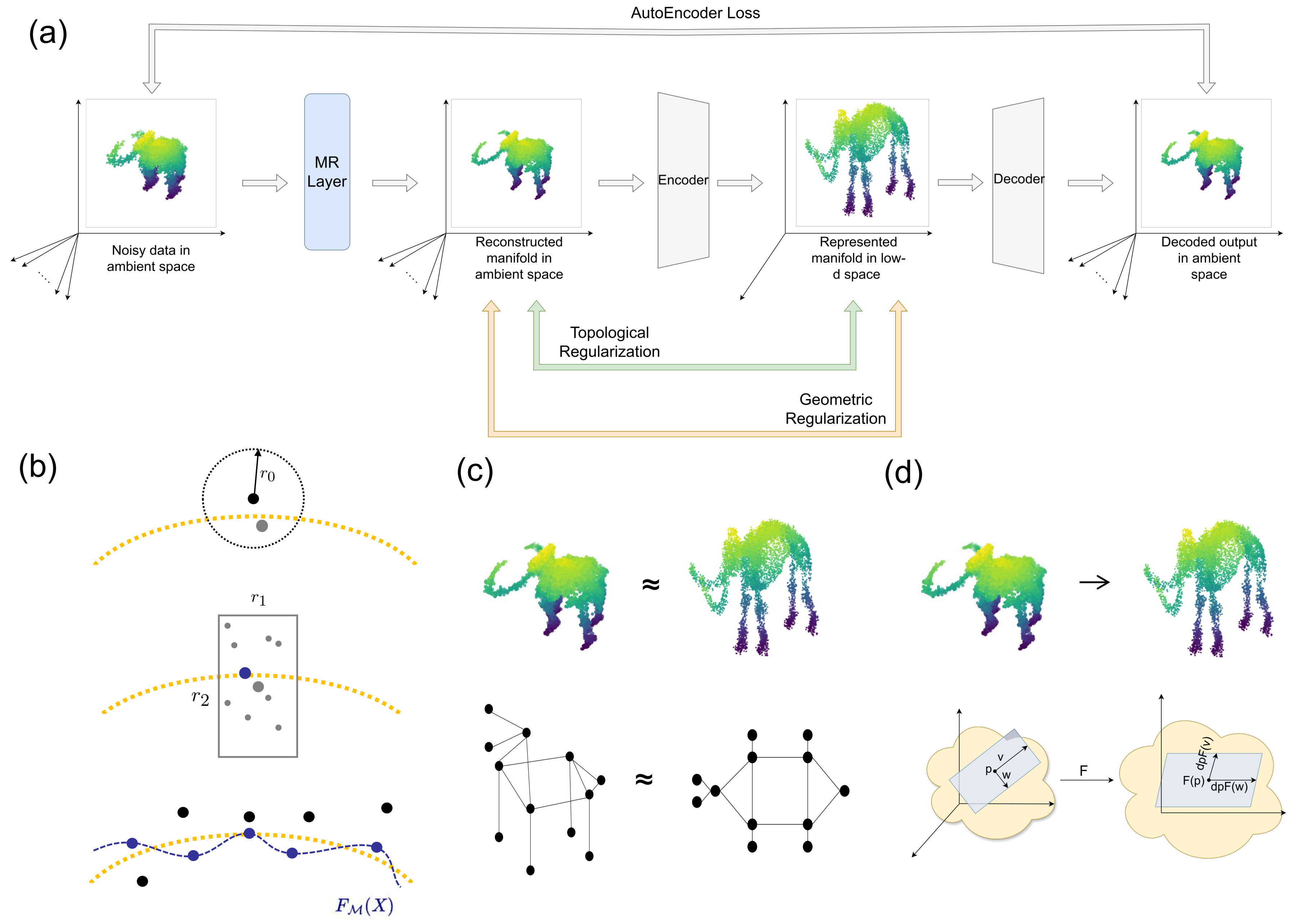}
    \caption{
    (a) An illustration of our model pipeline,
    (b) an illustration of our manifold reconstruction algorithm,
    (c) an illustration of the topological regularizer based on persistent homology,
    (d) an illustration of the geometric regularizer based on scaled isometry.
    }
    \label{fig: pipeline}
\end{figure}

\section{Preliminaries}

\subsection{Reconstructing the latent manifold}

\textit{Manifold fitting} is a challenging problem in manifold learning,
and it aims to reconstruct the smooth latent manifold from a data set that lies on or near it in the ambient Euclidean space.
Early manifold fitting approaches \cite{BGO09, CGW15} tackle noise-free samples with the Delaunay triangulation technique \cite{LS80}.
Recently, \citet{FMN16} validated the feasibility of manifold fitting under noise and proposed their own algorithms \cite{Fefferman1,Fefferman2,Fefferman3,Fefferman4}.
Genovese et. al. also provided a series of works from the perspective of minimax risk under Hausdorff distance \cite{Genovese1, Genovese2}.
An alternative approach is FlatNet by \citet{FlatNet} which flattens the noisy data points can reconstruct the manifold with linearized features.
The state-of-the-art manifold fitting algorithm proposed by \citet{YSLY23} features local contraction of noisy data points to an estimated direction.

\subsection{Topological regularization}

\hyperlink{Homology}{\emph{Homology}} is one of the most important topological invariants,
and is a natural target for classifying manifolds.
The prevailing method to obtain the homology of a point cloud is \hyperlink{Persistent Homology}{\emph{Persistent Homology}} \cite{PH},
which can be computed as a \hyperlink{Persistent diagram}{\emph{persistent diagram}}.

\citet{TopoAE} proposed a "Topological AutoEncoder" with a \emph{topological signature loss}, which computes the difference between persistent diagrams of the data set and the latent embedding,
thus encouraging the latent embedding to have the same topological or global structures as the original point cloud.
Besides the topological signature loss, there are also other methods to extract and compare topological properties of point clouds,
including Representation Topology Divergence (RTD) \cite{RTD,RTDAE} which modifies the computation of the persistent diagram, and Euler Characteristic Transform (ECT) \cite{ECT,DECT} which is a generalization of the Euler characteristics.

\subsection{Geometric regularization}

An \hyperlink{Isometry}{\emph{isometry}} between Riemannian manifolds preserves the Riemannian metric,
and thus is an equivalence up to geometric properties.
By encouraging the mapping to approximate an isometry,
one encourages the embedding not to be stretched or distorted,
but keeps the geometric or local structures near each point.

The approximation is done by optimizing the Jacobian matrix of the mapping.
\citet{GeomAE} proposed a "Geometric AutoEncoder" based on a generalized Jacobian determinant,
and \citet{IsoAE} achieved similar results by a coordinate-free relaxed distortion measure.
\citet{IsoLDM} promoted the relaxed distortion measure to Latent Diffusion Models.

\section{Methodology}

Due to the limited capacity and interest of the paper, we will omit most of the math derivations in this section. A more self-contained and detailed formulation of the math notions can be found in Appendix \ref{appendix: C} and references \cite{Lee, GeometricMeasureTheory, TopoAE}.

\subsection{The model pipeline}

The whole model is composed of a Manifold Reconstruction Layer (MRL), an Encoder and a Decoder,
as depicted in Figure \ref{fig: pipeline}.

The (mini-batched) input data is an $N\times D$ tensor $T_X\in\mathbb{R}^{N\times D}$,
where $N$ is the number of samples, 
and $D$ is the dimensionality of the Euclidean ambient space.
The MRL discovers the underlying manifold structures,
and contracts data points to the manifold,
resulting in $T_Y\in\mathbb{R}^{N\times D}$.
The Encoder compresses $T_Y$ to $T_Z\in\mathbb{R}^{N\times d}$,
where $d$ is the dimensionality of the representational space.
Finally, the Decoder tries to reconstruct $T_{\hat{X}}\in\mathbb{R}^{N\times D}$ from $T_Z$.
Here tensors $T_X$, $T_Y$, $T_Z$, $T_{\hat{X}}$ correspond to data sets $X\subset\mathcal{X}=\mathbb{R}^D$, $Y\subset\mathcal{Y}=\mathbb{R}^D$, $Z\subset\mathcal{Z}=\mathbb{R}^d$, $\hat{X}\subset\mathcal{X}=\mathbb{R}^D$.

The training loss function is the AutoEncoder loss between input data and decoded output, so the MRL is integrated in our training optimization. Note that we do not compute the loss between the reconstructed manifold and decoded output, otherwise the representation process will totally subject to the quality of the reconstructed manifold.

In addition, two training regularizers are computed between the reconstructed manifold and the embedding,
encouraging the preservation of local and global structures in the dimensionality reduction process.

\subsection{Assumptions on the data distribution}

In this paper, we treat our data as i.i.d. points residing on a latent manifold with extra Gaussian noise, and the manifold is regarded as a sub-manifold of Euclidean space. For further discussions and future work on different data settings, please refer to Section \ref{Sec: diss and future}.

Here we make several fundamental assumptions about our data:

\begin{assumption}
    The ambient space $\mathcal{X}$ is $D$-dimensional Euclidean space with the standard Euclidean \hyperlink{norm}{norm},
    namely,
    $\mathcal{X}=\mathbb{R}^D$.
\end{assumption}

\begin{assumption}
    The noise distribution $\phi_\sigma$ is a Gaussian distribution \hyperlink{support}{supported} on $\mathcal{X}$ with $0$ mean and standard deviation $\sigma$, whose density at point $\xi$ can be expressed as:
    \begin{equation}
        \phi_\sigma(\xi) = (\frac{1}{2\pi\sigma^2})^{\frac{D}{2}}\mathrm{exp}(-\frac{||\xi||^2_2}{2\sigma^2}).
    \end{equation}
\end{assumption}

\begin{assumption}
    The latent manifold $\mathcal{M}$ is a twice-differentiable and \hyperlink{compact}{compact} $L$-dimensional sub-manifold in $\mathcal{X}$.
\end{assumption}

\begin{assumption}
    The data distribution $\omega$ is a uniform distribution, with respect to the $L$-dimensional \hyperlink{Hausdorff measure}{Hausdorff measure},
    on $\mathcal{M}$.
\end{assumption}

Under these assumptions, we can draw a lemma about our data:

\begin{lemma} (Data formulation)
The set of data points $X\subset{\mathcal{X}}=\mathbb{R}^D$ can be written in the form of
    \begin{equation}
         x = x_{\mathcal{M}} + \xi,
        \ \forall x\in X,    
    \end{equation}
where $x_{\mathcal{M}} \sim \omega_{\mathcal{M}}$ and $\xi \sim \phi_\sigma$.
And the probability distribution can be written as a convolution
    \begin{equation}
        \nu(x)=\int_{\mathcal{M}}\omega_{\mathcal{M}}(t)\phi_\sigma(x-t)dt.
    \end{equation}
\end{lemma}

\subsection{The Manifold Reconstruction Layer}

The MRL has four hyperparameters: initial values of radii $r_0, r_1, r_2$,
and a smoothness parameter $k$,
affecting the computation of contraction directions and weights of the transformation matrix of each data point.
The values of $r_0, r_1, r_2$ are passed through back-propagation and optimized in the training process.

Following the algorithm provided in \cite{YSLY23},
we aim to develop an estimator $\hat{\mathcal{M}}$ for the latent manifold $\mathcal{M}$ using the data sample set $X\subset\mathcal{X}$.
Then for each $x\in X$, we employ a two-step procedure:
(i) we identify the contraction direction,
and (ii) we estimate the contracted point.

\textbf{Determine the contraction direction.}
Suppose the Gaussian noise distribution has standard deviation $\sigma$,
and $x^*=\text{argmin}_{x_\mathcal{M}\in\mathcal{M}}||x-x_\mathcal{M}||$.
We consider a $D$-dimensional ball $\mathcal{B}_D(x,r_0)$ with radius $r_0:=C_0\sigma$,
where $C_0$ is a given constant.
Then let $\mu^{\mathbb{B}}_x=\mathbb{E}_{T\sim\nu}(T|T\in\mathcal{B}_D(x,r_0))$,
we can estimate the direction of $x^* - x$ with $\mu^{\mathbb{B}}_x-x$.

To make the estimation smooth,
we compute $\mu^{\mathbb{B}}_x$ as $F(x)=\sum\alpha_i(x)x_i$,
and the weights $\alpha_i$'s are given as
\begin{equation}
    \tilde{\alpha}_i(x)=
    \begin{cases}
        (1-\frac{||x-x_i||^2_2}{r_0^2})^k, 
        &\ ||y-y_i||_2\le r_0; \\
        0, 
        & \ \text{otherwise},
    \end{cases}
    \ \tilde{\alpha}(x)=\sum_{i\in I_x}\tilde{\alpha}(x),
    \ \alpha_i(x)=\frac{\tilde{\alpha}_i(x)}{\tilde{\alpha}(x)},
\end{equation}
where $I_x$ is the set of indices for data points $x_i\in X\cap\mathcal{B}_D(x,r_0)$,
and $k>2$ is a fixed integer guaranteeing a twice-differentiable smoothness. 

\textbf{Determine the contracted point.}
Let $\tilde{U}$ be the projection matrix of $x$ onto the direction of $\mu_x^{\mathbb{B}}-x$.
Consider a cylinder region $\mathbb{V}_x=\mathcal{B}_{D-1}(x,r_1)\times\mathcal{B}_1(x,r_2)$,
where the second ball is an open interval in the direction of $\mu_x^{\mathbb{B}}-x$,
and the first ball is in the complement of it in $\mathcal{X}=\mathbb{R}^D$,
with $r_1 := C_1\sigma$ and $r_2 := C_2\sigma\sqrt{\log(1/\sigma)}$.
Then the contracted version of $x$ is denoted as $\mu^{\mathbb{V}}_x=x+\tilde{U}\mathbb{E}_{T\sim\nu}(T-x|T\in\mathbb{V}_x)$.

For computational cause, we let $\hat{U}=\frac{(F(x)-x)(F(x)-x)^T}{||F(x)-x||^2_2}$.
For a data point $x_j\in\mathbb{V}_x$ ($j\in J_x$),
we define $u_j=\hat{U}(x_j-x)$, $v_j=x_j-x-u_i$.
Then we can construct a smooth map $G(x)=\sum\beta_j(x)x_j$ with weights given as
\begin{equation}
    \label{FM}
    \begin{aligned}
        &w_u(u_j)=
        \begin{cases}
            1, 
            &\ ||u_j||_2\le \frac{r_2}{2}; \\
            [1-(\frac{2||u_j||_2-r_2}{r_2})^2]^k, 
            & \ ||u_j||_2\in(\frac{r_2}{2},r_2); \\
            0, 
            & \ \text{otherwise},
        \end{cases} \\
        &w_v(v_j)=
        \begin{cases}
            (1-\frac{||v_j||^2_2}{r_1^2})^k, 
            &\ ||v_j||_2\le r_1; \\
            0, 
            & \ \text{otherwise},
        \end{cases} \\
        &\beta_j(x)=w_u(u_j)w_v(v_j),\ \tilde{\beta}(x)=\sum_{j\in J_x}\tilde{\beta}_j(x),\ \beta_j(x)=\frac{\tilde{\beta}_j(x)}{\tilde{\beta}(x)}.
    \end{aligned}
\end{equation}

Finally, we get a function $F_{\mathcal{M}}:\mathcal{X}\to\mathcal{X}$ where $F_{\mathcal{M}}(x) := y\in Y$ is our estimated value of $x^*$,
thus a reconstructed data point representing $x_{\mathcal{M}}$.
An illustration of the process is shown in Figure \ref{fig: pipeline}.

\subsection{The topological regularizer}

\textbf{Persistent diagrams.}
Persistent Homology (PH) is computed first by constructing a \emph{Vietoris-Rips complex}.
Given a distance scale $\epsilon$,
the Vietoris-Rips complex of $X$,
denoted by $\mathfrak{K}_\epsilon(X)$,
contains all simplices (i.e. subsets) of $X$ whose elements have mutual distances smaller than or equal to $\epsilon$.
Vietoris-Rips complexes form a \emph{filtration},
i.e. $\mathfrak{K}_{\epsilon_1} \subseteq \mathfrak{K}_{\epsilon_2}$ for $\epsilon_1\le\epsilon_2$.
With this quality, we can track changes in homology groups of Vietoris-Rips complexes as $\epsilon$ increases from $0$.
In the increasing process,
the connectivity among points in $X$ changes accordingly,
and each connectivity state has "birth and death times" denoted by two values $(a,b)$ of $\epsilon$.
Thus we can track the creation and destruction of $n$-dimensional topological features in a diagram $\mathcal{D}_n$,
called the $n$-th \hyperlink{Persistent diagram}{\emph{persistent diagram}}.

\textbf{The topological signature loss.}
In our method, the PH information of $X$ is stored in a tuple $(\mathcal{D}^X:=\{\mathcal{D}_0,\mathcal{D}_1,\cdots\}, \pi^X:=\{\pi_0,\pi_1,\cdots\})$,
where the first component is the persistent diagrams, and the second component is the \emph{persistent pairings} composing of pairs of edges $(e,e')$ that each create or destroy a topological feature $(a,b)\in\mathcal{D}_n$ when they are newly connected as $\epsilon$ increases.

The values of the persistent diagram can be retrieved by subsetting the mutual distance matrix $\mathbf{A}^X$ of points in $X$ with the edge indices provided by the persistent pairings,
written as
\begin{equation}
    \mathbf{A}^X[\pi^X]=\{(|e|, |e'|),\forall (e,e')\in\pi^X\}\in\mathbb{R}^{|\pi^X|}.
\end{equation}
For the manifold $Y$ and embedding $Z$,
we define a topological signature loss of two terms,
each denoting the loss measuring the changes of topological features in one of the two Vietors-Rips complexes compared to those in the other:
\begin{equation}
\begin{aligned}
        \mathcal{L}_{topo} = &\mathcal{L}_{Y\to Z} + \mathcal{L}_{Z\to Y} \\
                           = &\frac{1}{2}||\mathbf{A}^Y[\pi^Y] - \mathbf{A}^Z[\pi^Y]||^2 + \frac{1}{2}||\mathbf{A}^Z[\pi^Z] - \mathbf{A}^Y[\pi^Z]||^2.
\end{aligned}
\end{equation}

A lower topological signature loss means the topological features are better shared between the manifold $Y$ and embedding $Z$.

\begin{definition}
    The topological regularizer is given by
    \begin{equation}
        \mathcal{L}_{topo} = \mathrm{TopoSig}(Y, Z, \pi_Y, \pi_Z) =
        \frac{1}{2}||\mathbf{A}^Y[\pi^Y] - \mathbf{A}^Z[\pi^Y]||^2 + \frac{1}{2}||\mathbf{A}^Z[\pi^Z] - \mathbf{A}^Y[\pi^Z]||^2,
    \end{equation}
    where $A$ is a distance matrix,
    $\pi$ is a Vietoris-Rips complex,
    and $A[\pi]$ is a selection of distance entries corresponding to topologically relevant edges in the Vietoris-Rips complex.
\end{definition}

\subsection{The geometric regularizer}

\textbf{Scaled isometry.}
As its name suggests, the relaxed distortion measure indicates the degree of distortion of a mapping compared to a \hyperlink{Isometry}{\emph{scaled isometry}}.
Let $\mathcal{M}$ be a Riemannian manifold of dimension $m$ with local coordinates $x\in\mathbb{R}^m$ and Riemannian metric $G(x)\in\mathbb{R}^{m\times m}$,
and $\mathcal{N}$ be a Riemannian manifold of dimension $n$ with $z\in\mathbb{R}^n$ and $H(z)\in\mathbb{R}^{n\times n}$.
Let $f:\mathcal{M}\to\mathcal{N}$, then its differential is denoted by the Jacobian matrix $J_f(x)=\frac{\partial f}{\partial x}(x)\in\mathbb{R}^{n\times m}$.
$f$ is said to be a scaled isometry if it meets $G(x)=CJ_f(x)^TH(f(x))J_f(x),\ \forall x\in\mathbb{R}^m$, for a constant $C$.
An illustration is in Figure \ref{fig: pipeline}.

\textbf{Relaxed distortion measure.}
At a point $x\in\mathcal{M}$, we consider the characteristic values of the \hyperlink{pullback metric}{\emph{pullback metric}} $J_f^THJ_f$ relative to $G$, 
i.e. the $m$ eigenvalues $\lambda_1, \cdots,\lambda_m$ of $J_f(x)^TH(f(x))J_f(x)^TG^{-1}(x)$.
These eigenvalues are invariant under coordinate transformations,
and let $S(\lambda_1,\cdots,\lambda_m)$ be any symmetric function,
the integral $\mathcal{I}_S(f):=\int_{\mathcal{M}}S(\lambda_1(x),\cdots,\lambda_m(x))d\mu(x)$ is also coordinate-invariant.
Here $\mu$ is a positive measure on $\mathcal{M}$.

Here we design a family of coordinate-invariant functionals $\mathcal{F}$ that measures how far $f:\mathcal{M}\to\mathcal{N}$ is from being a scaled isometry over the support of $\mu$:
\begin{equation}
    \mathcal{F}(f)=\int_{\mathcal{M}}\sum^m_{i=1}h(\frac{\lambda_i(x)}{\int_{\mathcal{M}}S(\lambda_1(x),\cdots,\lambda_m(x))d\mu(x)})d\mu(x),
\end{equation}
where $h$ is some convex function and $S$ is some symmetric function.
Further, if we have $\mu$ \textcolor{red}{as} a finite measure, $S(k\lambda_1,\cdots,k\lambda_m)=kS(\lambda_1,\cdots,\lambda_m)$, and $S(1,\cdots,1)=\frac{1}{\mu(\mathcal{M})}$,
the functional satisfies the following conditions:
\begin{enumerate}
    \item $\mathcal{F}(f)\ge0$;
    \item $\mathcal{F}(f)=0$ if and only if $\lambda_i(x)=c$ for $\forall i$, $\forall x\in\text{Supp}(\mu)$, and for some $c>0$;
    \item $\mathcal{F}(f)=\mathcal{F}(g)$ if $J^T_fHJ_f=CJ^T_gHT_g$ for $\forall x\in\text{Supp}(\mu)$, and for some $C>0$.
\end{enumerate}
With this given functional, we can calculate and minimize the relaxed distortion measure of our encoder function.

\begin{definition}
    The geometric regularizer is given by
    \begin{equation}
        \mathcal{L}_{geom} = \mathcal{F}(f_\theta;P_\phi)
        = \mathbb{E}_{y\sim P_\phi}[\sum_{i=1}^D(\frac{\lambda_i(y)}{\mathbb{E}_{y\sim P_\phi}[\sum_i\frac{\lambda_i(y)}{D}]} - 1)^2]
        = D^2\frac{\mathbb{E}_{y\sim P_\phi}[\mathrm{Tr}(H_\theta^2(y))]}{\mathbb{E}_{y\sim P_\phi}[\mathrm{Tr}(H_\theta(y))]^2}-D,
    \end{equation}
    where $f_\theta: \mathbb{R}^D \to \mathbb{R}^d$, $y\mapsto z$ is the encoder function,
    $P_\phi$ is the distribution of $Y$ in $\mathbb{R}^D$: $y\sim P_\phi$,
    $\lambda_i(y)$ are eigenvalues of $H_\theta(y) = J_{f_\theta}^T(y)H(f_\theta(y))J_{f_\theta}(y)$.
    Here $H(z)$ is the metric assigned to some data point $z\in Z$ in the embedding,
    and $H_\theta(y)$ is the pullback metric assigned to the corresponding $y\in Y$ in the reconstructed manifold.

\end{definition}

\subsection{The total loss function}
\begin{definition}
    The total loss function is given by
    \begin{equation}
        \mathcal{L} = \lambda_{AE}\mathcal{L}_{AE} (X,\hat{X})
        + \lambda_{topo}\mathcal{L}_{topo} (Y,Z)
        + \lambda_{geom}\mathcal{L}_{geom} (Y,Z),
    \end{equation}
    where $\lambda_{AE}$, $\lambda_{topo}$, and $\lambda_{geom}$ are predefined hyperparameters,
    and $X$, $Y$, $Z$, $\hat{X}$ denotes the input noisy data set, the reconstructed manifold, the low-dimensional embedding, and the AutoEncoder output result, respectively.
\end{definition}

\section{Experiments}

\begin{figure}[t]
    \centering
    \includegraphics[width=1\linewidth]{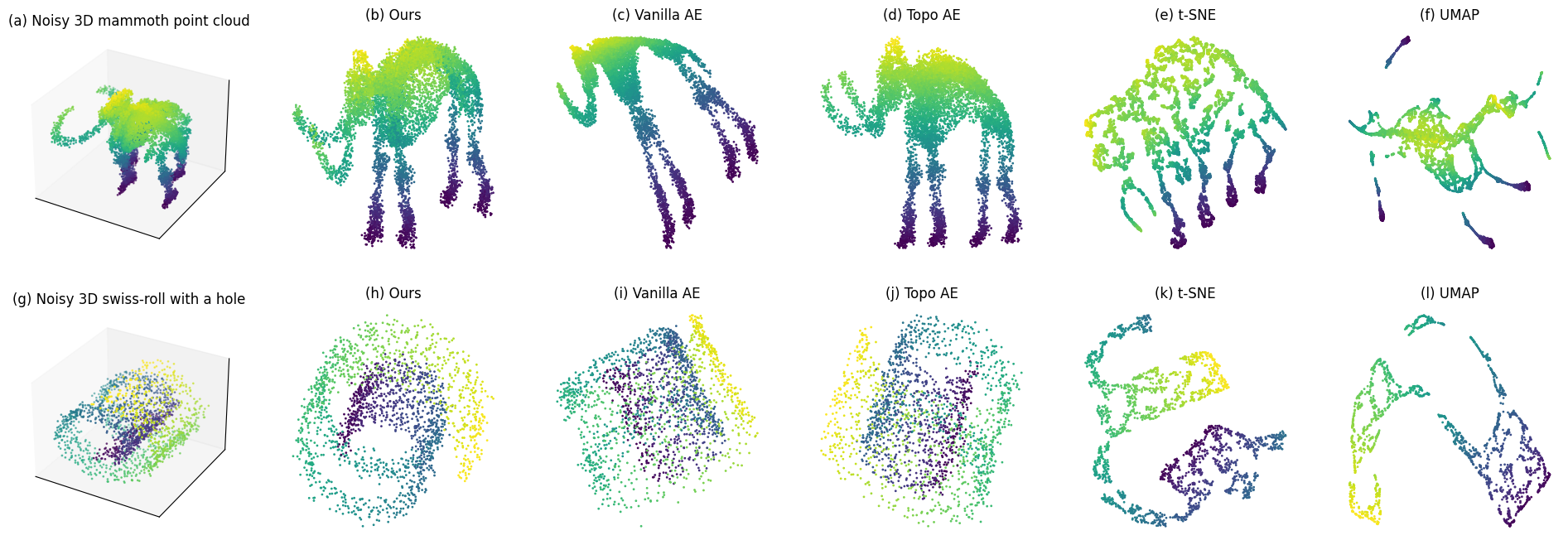}
    \caption{
            Dimensionality reduction of 3D point cloud data to 2D space.
             Our proposed method best preserves global and local structures of the latent manifolds underlying noisy data.
             }
    \label{fig:overview}
\end{figure}

\subsection{Datasets}

In this paper, we implement our method on Euclidean point cloud datasets to validate our approach for manifold representation learning. 
Datasets include a noisy 3D Swiss roll manifold with a hole, produced from the dataset in Scikit-learn \cite{scikit-learn}; a noisy 3D mammoth dataset, produced from the mammoth dataset shown in the "Understanding UMAP" project \cite{Mammoth}; a 100-dimensional spheres dataset from the "Topological AutoEncoder" paper \cite{TopoAE}; and a subset of the 3D point cloud dataset PartNet \cite{PartNet}.
A detailed demonstration of the datasets can be found in Appendix \ref{Appendix: datasets}.

\subsection{Baselines}

We train the following baseline manifold learning methods on the datasets, and compare their performance with our method by showing the 2-dimensional embedding shapes, and evaluating with dimensionality reduction benchmarking metrics adopted from \cite{TopoAE,GeomAE}.

A vanilla AutoEncoder and a Topological AutoEncoder are trained with the same input data, model structure, loss function, and learning rate as our model.
Standard PCA, t-SNE, and UMAP models fit-transform the same data as our model,
all with default parameters.

\subsection{Architecture and training}

For 3D point cloud data, we build our AE model with a "3-2-2" encoder and a "2-2-3" decoder, with GELU \cite{GELU} activations. For spheres data in the 101-D space, the model has a "101-64-32-16-8-4-2" encoder and a "2-4-8-16-32-64-101" decoder.
MRL hyperparameters are $(r_0,r_1,r_2,k)=(1.0,0.01,1.0,3)$ for all experiments.
The loss term weights $(\lambda_{AE},\lambda_{topo},\lambda_{geom})$ are $(1,1,5)$ for the Swiss roll dataset, $(1,0.5,0.5)$ for the mammoth dataset,  $(1,0.01,0.01)$ for the PartNet dataset, and $(1,1,1)$ for the spheres dataset.
The AutoEncoder loss is set as the MSE loss between the model input and output.
A collection of the hyperparameter settings and training details can be found in Appendix \ref{Appendix: training}.

\subsection{Visualization}

\textbf{Swiss roll and mammoth datasets.}
The dimensionality reduction results are shown in Figure \ref{fig:overview}.
We can see that Vanilla AE reduces the manifolds as a whole to 2D space, but the embeddings lost many details (two tusks of the mammoth, the "roll" shape, and the hole on the Swiss roll), and are heavily distorted.
The Topo AE embeddings keep their global shape, but their local structures are ignored, as both embeddings are "folded" into 2D space,
producing non-stereoscopic self-covering embeddings.
T-SNE and UMAP embeddings primarily focus on local structural information, while the manifolds are torn apart with unrecognizable global shapes.
Our proposed method keeps the most local and global shape details, and preserves the proportions of the manifold not distorted or torn apart,  
thus leaving a stereoscopic representation of the point cloud.

\begin{figure}[h]
    \centering
    \includegraphics[width=1\linewidth]{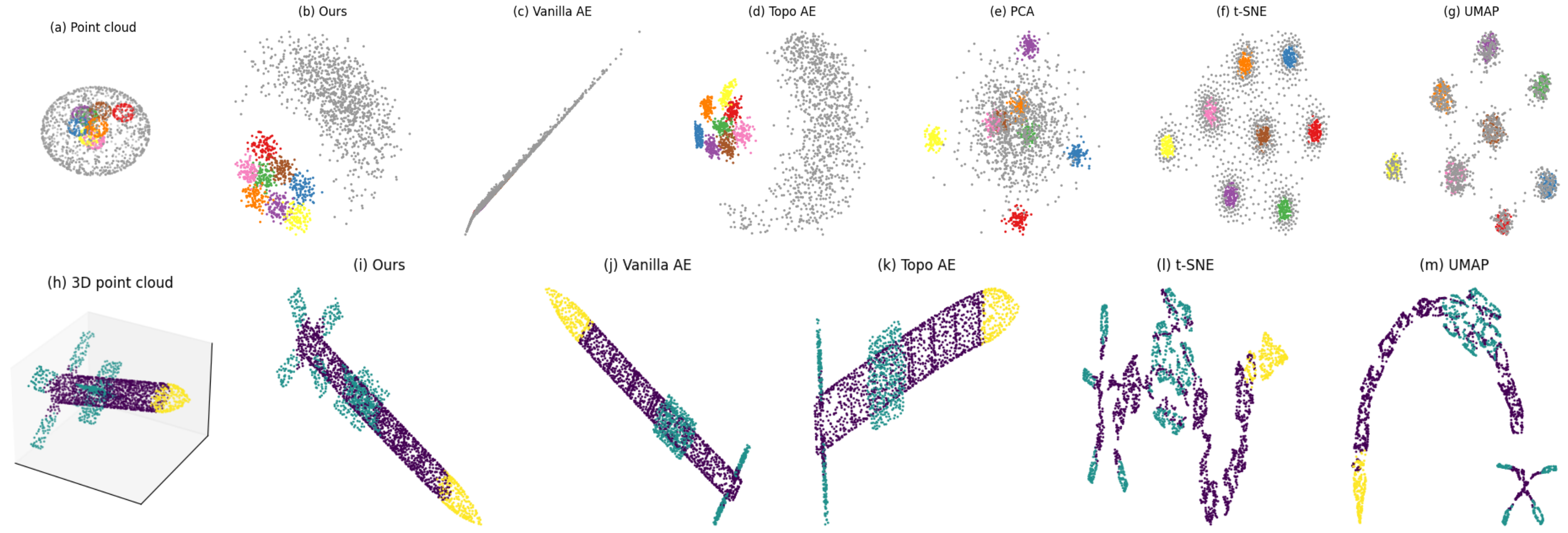}
    \caption{Dimensionality reduction of 100-D spheres, and "rocket" object from PartNet dataset.}
    \label{fig:shperes}
\end{figure}

\textbf{Spheres dataset.}
The dimensionality reduction results are shown in Figure \ref{fig:shperes}.
The dataset is composed of nine 100-dimensional spheres in 101-dimensional Euclidean space,
where eight small spheres are located inside a large sphere,
as demonstrated in panel (a).

Similar to the above results, only our method and Topo AE can separate the large surrounding sphere from small spheres,
and our method further captures the round shape and a shared radius of inner spheres. 
The Vanilla AE produces incomprehensible embeddings, PCA just projects high-dimensional data to 2D space, and t-SNE and UMAP mix the large shpere with clusters of small spheres.

\textbf{PartNet dataset.}
The dataset is composed of 12 distinct 3D point clouds, each indicating a real-life object.
The points on an object are labeled by a ground truth semantic segmentation,
where different colors indicate different "parts" of the object.
Dimensionality reduction results of the "rocket" object are in Figure \ref{fig:shperes},
and results of the whole dataset can be found in Figure \hyperlink{Figure S1}{S1}.

Similar to the above results, our method best preserves the local and global shapes of different objects,
whereas the baseline models produce flattened, distorted, or fractured embedding shapes.
Worth noting are the correct shapes of complex structures such as the chair, the motorbike, and the pistol,
and how our model keeps the triangular legs of the table without tearing it apart.

\subsection{Evaluation}

We adopted the evaluation metrics from \cite{TopoAE,GeomAE} for measuring the similarity of local and global structures between the original point cloud and the embedding.
The metrics are separated into two groups: $\mathrm{KL_{0.1}}$, kNN and Trust are local metrics, while $\mathrm{KL_{100}}$, RMSE and Spear are global metrics.
For a detailed introduction of the quality measures, please refer to Appendix \ref{Appendix: metrics}.

In Table \ref{tab: metrics}, we can see that our method keeps a balance between local and global structures.
Worth noting is that in most of the metrics rankings, our model outperforms Vanilla AE.
The full table for PartNet evaluation metrics can be found in Table \hyperlink{Table S1}{S1}.

\begin{table}[t]
\caption{
         Local and global metrics for our method and manifold learning baselines.
         PartNet scores are averaged over 12 objects.
         \textbf{Bold} values are winners,
         and \underline{underlined} are runner-ups.
        }
\label{tab: metrics}
\resizebox{\textwidth}{!}
{
    \begin{tabular}{@{}cccccccc@{}}
    \toprule
    \multirow{2}{*}{Dataset} & \multirow{2}{*}{Method} & \multicolumn{3}{c}{Local} & \multicolumn{3}{c}{Global} \\ \cmidrule(l){3-8}
                                &            & $\mathrm{KL_{0.1}}(\downarrow)$ & kNN$(\uparrow)$ & Trust$(\uparrow)$ & $\mathrm{KL_{100}}(\downarrow)$ & RMSE$(\downarrow)$ & Spear$(\uparrow)$ \\ \midrule

    \multirow{5}{*}{Swiss roll} & Ours       & $\mathbf{2.19\times10^{-2}}$    & $4.69\times10^{-1}$             & $8.98\times10^{-1}$      
                                             & $\underline{1.12\times10^{-7}}$ & $\mathbf{2.55\times10^{-1}}$    & $\mathbf{7.60\times10^{-1}}$      \\
                                             
                                & Vanilla AE & $3.07\times10^{-2}$             & $4.17\times10^{-1}$             & $8.84\times10^{-1}$      
                                             & $1.13\times10^{-7}$             & $\underline{2.56\times10^{-1}}$ & $7.41\times10^{-1}$     \\
                                             
                                & Topo AE    & $\underline{2.90\times10^{-2}}$ & $5.34\times10^{-1}$             & $8.79\times10^{-1}$     
                                             & $\mathbf{1.11\times10^{-7}}$    & $4.07\times10^{-1}$             & $\underline{7.59\times10^{-1}}$      \\
                                             
                                & t-SNE      & $2.95\times10^{-2}$             & $\mathbf{8.27\times10^{-1}}$    & $\mathbf{9.98\times10^{-1}}$     
                                             & $2.68\times10^{-7}$             & $503\times10^{-1}$                & $4.79\times10^{-1}$      \\
                                             
                                & UMAP       & $3.20\times10^{-2}$             & $\underline{7.80\times10^{-1}}$ & $\underline{9.96\times10^{-1}}$      
                                             & $4.77\times10^{-7}$             & $114\times10^{-1}$                & $2.50\times10^{-1}$      \\ \midrule

    \multirow{5}{*}{Mammoth}    & Ours       & $\mathbf{2.15\times10^{-3}}$    & $4.93\times10^{-1}$             & $9.61\times10^{-1}$      
                                             & $\mathbf{1.03\times10^{-8}}$    & $1.63\times10^{-1}$             & $\mathbf{9.76\times10^{-1}}$      \\
    
                                & Vanilla AE & $21.6\times10^{-3}$             & $2.05\times10^{-1}$             & $9.38\times10^{-1}$      
                                             & $4.25\times10^{-8}$             & $\underline{1.38\times10^{-1}}$ & $7.62\times10^{-1}$      \\
                                
                                & Topo AE    & $\underline{2.94\times10^{-3}}$ & $4.96\times10^{-1}$             & $9.65\times10^{-1}$      
                                             & $\underline{1.75\times10^{-8}}$ & $\mathbf{1.21\times10^{-1}}$    & $\underline{9.64\times10^{-1}}$      \\
                                
                                & t-SNE      & $27.5\times10^{-3}$             & $\mathbf{7.23\times10^{-1}}$    & $\mathbf{9.98\times10^{-1}}$     
                                             & $12.5\times10^{-8}$             & $871\times10^{-1}$              & $8.13\times10^{-1}$      \\
                                
                                & UMAP       & $10.9\times10^{-3}$             & $\underline{6.62\times10^{-1}}$ & $\underline{9.97\times10^{-1}}$      
                                             & $8.52\times10^{-8}$             & $149\times10^{-1}$              & $8.51\times10^{-1}$      \\ \midrule

    \multirow{5}{*}{PartNet}    & Ours       & $\mathbf{8.18\times10^{-2}}$    & $5.40\times10^{-1}$             & $9.24\times10^{-1}$      
                                             & $\mathbf{6.02\times10^{-8}}$    & $\underline{3.44\times10^{-1}}$ & $\mathbf{8.81\times10^{-1}}$      \\
    
                                & Vanilla AE & $18.8\times10^{-2}$             & $5.08\times10^{-1}$             & $9.25\times10^{-1}$      
                                             & $16.8\times10^{-8}$             & $10.1\times10^{-1}$             & $8.24\times10^{-1}$      \\
                                
                                & Topo AE    & $18.3\times10^{-2}$             & $5.43\times10^{-1}$             & $9.32\times10^{-1}$      
                                             & $15.6\times10^{-8}$             & $\mathbf{2.21\times10^{-1}}$    & $8.61\times10^{-1}$      \\

                                & t-SNE      & $16.5\times10^{-2}$             & $\mathbf{7.85\times10^{-1}}$    & $\mathbf{9.96\times10^{-1}}$      
                                             & $\underline{11.8\times10^{-8}}$ & $526\times10^{-1}$              & $\underline{8.76\times10^{-1}}$      \\
                                
                                & UMAP       & $\underline{16.0\times10^{-2}}$ & $\underline{7.50\times10^{-1}}$ & $\underline{9.94\times10^{-1}}$      
                                             & $13.0\times10^{-8}$             & $93.8\times10^{-1}$             & $8.58\times10^{-1}$      \\ \midrule

    \multirow{6}{*}{Spheres}    & Ours       & $\underline{4.78\times10^{-1}}$ & $1.76\times10^{-1}$             & $5.78\times10^{-1}$      
                                             & $10.1\times10^{-7}$             & $2.81\times10^{1}$              & $-1.85\times10^{-2}$      \\
    
                                & Vanilla AE & $4.90\times10^{-1}$             & $2.23\times10^{-1}$             & $6.18\times10^{-1}$      
                                             & $\underline{9.28\times10^{-7}}$ & $6.48\times10^{1}$              & $\mathbf{27.7\times10^{-2}}$      \\
                                
                                & Topo AE    & $\mathbf{3.97\times10^{-1}}$    & $1.53\times10^{-1}$             & $5.64\times10^{-1}$      
                                             & $\mathbf{9.12\times10^{-7}}$    & $2.82\times10^{1}$              & $5.21\times10^{-2}$      \\
                                
                                & PCA        & $7.49\times10^{-1}$             & $2.16\times10^{-1}$             & $6.28\times10^{-1}$      
                                             & $15.4\times10^{-7}$             & $\underline{2.31\times10^{1}}$  & $-3.45\times10^{-2}$      \\
                                
                                & t-SNE      & $5.95\times10^{-1}$             & $\mathbf{3.50\times10^{-1}}$    & $\mathbf{7.33\times10^{-1}}$      
                                             & $12.7\times10^{-7}$             & $\mathbf{1.43\times10^{1}}$     & $\underline{7.38\times10^{-2}}$      \\
                                
                                & UMAP       & $6.17\times10^{-1}$             & $\underline{2.59\times10^{-1}}$ & $\underline{6.83\times10^{-1}}$      
                                             & $14.2\times10^{-7}$             & $2.40\times10^{1}$              & $4.26\times10^{-2}$      \\ \bottomrule
    \end{tabular}
}
\end{table}

\section{Ablation study}

To validate the necessity of our model ingredients,
we perform an ablation study on the model pipeline with the Swiss roll dataset.
We compared the performance of ablated models as 
\begin{enumerate}
    \item "Final model": Manifold Reconstruction Layer + AutoEncoder + two regularizers;
    \item "Topo AE": AutoEncoder + Topological Regularizer;
    \item "Geom AE": AutoEncoder + Geometric Regularizer;
    \item "Topo-geom AE": AutoEncoder + two regularizers;
    \item "MR AE": Manifold Reconstruction Layer + AutoEncoder.
\end{enumerate}
Visualizations and evaluation metrics are shown in Figure \hyperlink{Figure S2}{S2} and Table \hyperlink{Table S2}{S2}.

Note that as the figure depicts, our model has a cleaner reconstructed manifold than that of MR AE.
In the "Point Cloud vs Manifold" group of the table, our model beats MR AE in all the metrics,
indicating that the representation component (the AutoEncoder) improves the performance of the Manifold Reconstruction Layer during training.
Our model's performance in "Manifold vs Embedding" also exceeds Topo-geom AE's in "Point Cloud vs Embedding" in four out of six metrics,
which means the reconstructed manifold discovers local and global properties out of the noisy point cloud,
which can be more easily carried to low-dimensional space under the same dimensionality reduction method.

In summary, the manifold reconstruction and representation regularization components are both virtuous in our implementation,
and they promote each other's performance during the training optimization,
"clearer answers are easier to be written down with the same pen, and clearer writing can improve scores earned by the same answer",
as we previously claimed.

\section{Limitations and future directions}
\label{Sec: diss and future}

\textbf{Supervision of Manifold Reconstruction.}
The Manifold Reconstruction process alters the data, while the Autoencoder faithfully embeds its results.
In our implementation, we improve the MRL's performance through the encode-decode feedback,
but this may not be a sufficient supervision,
and the learning process could be unstable.
In the future, we aim to come up with better ways to supervise the manifold reconstruction process.

\textbf{Non-Euclidean data.}
When dealing with data types other than point clouds in Euclidean space, we need to modify our manifold reconstruction method:
if the coordinate axes are correlated, we need to first orthogonalize them;
when the data are sequential, we need to integrate the sequence information into our distance function;
and if we have external labels for data points, we should learn a distance metric with them.

\newpage

\bibliography{reference}

\newpage

\appendix

\section{Extended figures and tables}
\setcounter{figure}{0}
\renewcommand{\thefigure}{S\arabic{figure}}

\setcounter{table}{0}
\renewcommand{\thetable}{S\arabic{table}}

\hypertarget{Figure S1}{}
\begin{figure}[h]
    \centering
    \includegraphics[width=1\linewidth]{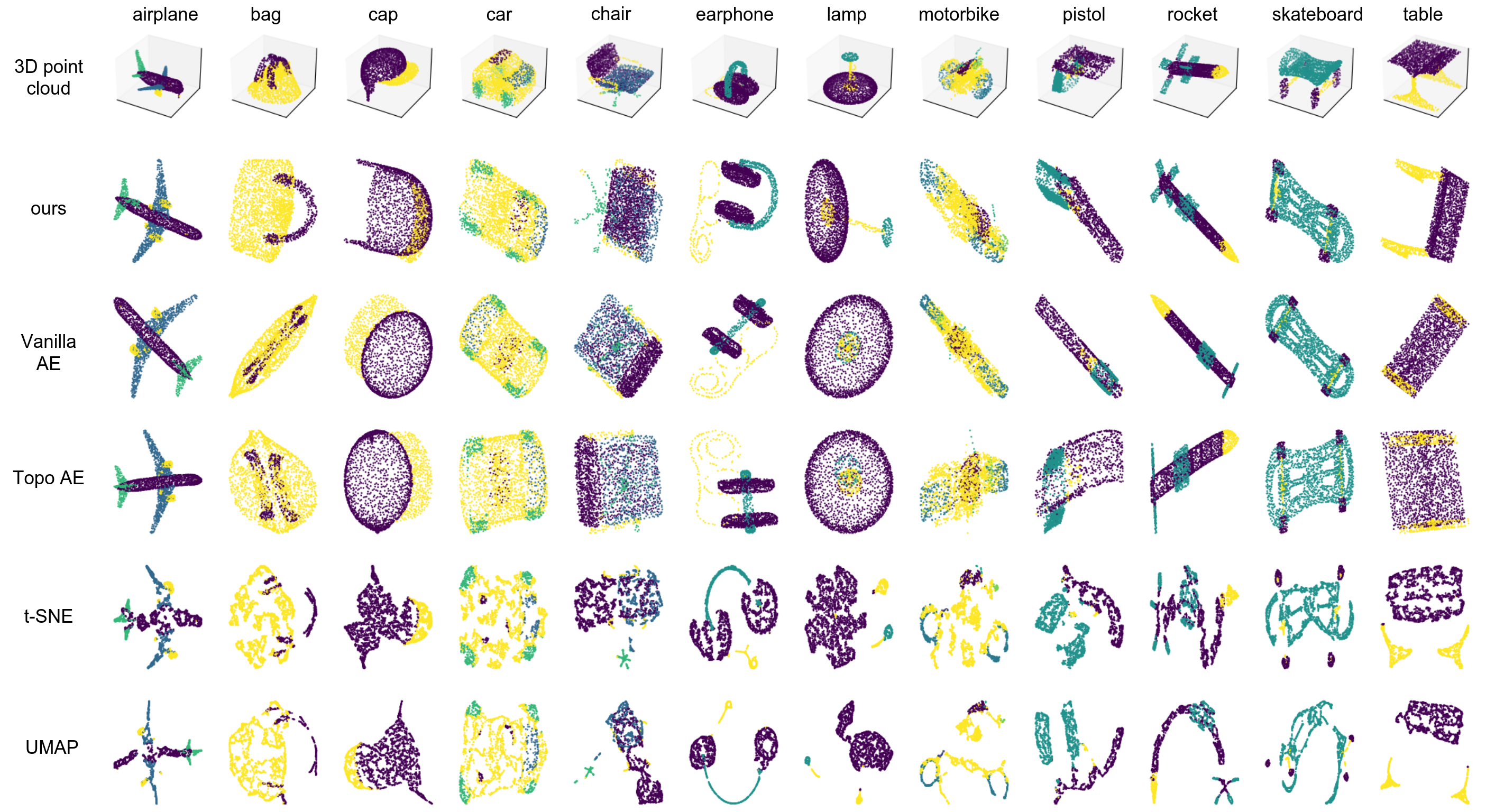}
    \caption{Dimensionality reduction results of a 12-object subset of the PartNet dataset.
             The rows are the original 3D point cloud of the object,
             and 5 different manifold learning methods to be compared.
             The columns are 12 different objects,
             with ground-truth semantic segmentation of different parts.}
\end{figure}

\textbf{Comments.}
Note that our embeddings are the most "stereoscopic",
as they keep a balance between the global structure and local rigidity.
The global shapes and the proportions of each part are preserved and presented in 2D space,
in a way that they stay in a whole, but do not cover or interrupt each other (with the best effort).

In the meantime,
Vanilla AE and Topo AE try to preserve the whole global structure,
however, the embeddings are often flattened or distorted, as if all the structural information is "shoved" into low-dimensional space.

While t-SNE and UMAP focus too much on local similarity,
tearing apart the points that are relatively distant but actually connected,
resulting in fractured pieces without recognizable global shapes. 


\hypertarget{Table S1}{}
\begin{table}[]
\centering
\caption{
         Local and global metrics for 12 PartNet objects.
        }
\resizebox{0.94\textwidth}{!}
{
    \begin{tabular}{@{}cccccccc@{}}
    \toprule
    \multirow{2}{*}{Object} & \multirow{2}{*}{Method} & \multicolumn{3}{c}{Local} & \multicolumn{3}{c}{Global} \\ \cmidrule(l){3-8}
                                &            & $\mathrm{KL_{0.1}}(\downarrow)$ & kNN$(\uparrow)$ & Trust$(\uparrow)$ & $\mathrm{KL_{100}}(\downarrow)$ & RMSE$(\downarrow)$ & Spear$(\uparrow)$ \\ \midrule

    \multirow{5}{*}{Airplane}   & Ours       & $7.81\times10^{-3}$             & $6.11\times10^{-1}$             & $9.70\times10^{-1}$      
                                             & $51.6\times10^{-9}$             & $3.66\times10^{-1}$             & $9.23\times10^{-1}$      \\
                                             
                                & Vanilla AE & $3.23\times10^{-3}$             & $6.98\times10^{-1}$             & $9.91\times10^{-1}$      
                                             & $31.6\times10^{-9}$             & $11.2\times10^{-1}$             & $9.37\times10^{-1}$     \\
                                             
                                & Topo AE    & $8.28\times10^{-3}$             & $7.28\times10^{-1}$             & $9.92\times10^{-1}$     
                                             & $4.89\times10^{-9}$             & $2.08\times10^{-1}$             & $9.94\times10^{-1}$      \\
                                             
                                & t-SNE      & $8.36\times10^{-3}$             & $7.86\times10^{-1}$             & $9.97\times10^{-1}$     
                                             & $37.9\times10^{-9}$             & $526\times10^{-1}$              & $9.27\times10^{-1}$      \\
                                             
                                & UMAP       & $4.23\times10^{-3}$             & $7.31\times10^{-1}$             & $9.95\times10^{-1}$      
                                             & $22.3\times10^{-9}$             & $102\times10^{-1}$              & $9.55\times10^{-1}$      \\ \midrule

    \multirow{5}{*}{Bag}        & Ours       & $4.46\times10^{-3}$             & $6.53\times10^{-1}$             & $9.77\times10^{-1}$      
                                             & $3.05\times10^{-8}$             & $3.55\times10^{-1}$             & $9.65\times10^{-1}$      \\
                                             
                                & Vanilla AE & $29.7\times10^{-3}$             & $3.42\times10^{-1}$             & $8.51\times10^{-1}$      
                                             & $22.6\times10^{-8}$             & $11.9\times10^{-1}$             & $6.27\times10^{-1}$     \\
                                             
                                & Topo AE    & $27.1\times10^{-3}$             & $3.90\times10^{-1}$             & $8.74\times10^{-1}$     
                                             & $21.8\times10^{-8}$             & $3.51\times10^{-1}$             & $6.44\times10^{-1}$      \\
                                             
                                & t-SNE      & $14.8\times10^{-3}$             & $7.73\times10^{-1}$             & $9.95\times10^{-1}$     
                                             & $7.58\times10^{-8}$             & $471\times10^{-1}$              & $9.17\times10^{-1}$      \\
                                             
                                & UMAP       & $28.1\times10^{-3}$             & $7.66\times10^{-1}$             & $9.96\times10^{-1}$      
                                             & $14.2\times10^{-8}$             & $72.1\times10^{-1}$             & $8.54\times10^{-1}$      \\ \midrule

    \multirow{5}{*}{Cap}        & Ours       & $121\times10^{-3}$              & $5.56\times10^{-1}$             & $9.03\times10^{-1}$      
                                             & $9.04\times10^{-8}$             & $3.47\times10^{-1}$             & $8.08\times10^{-1}$      \\
                                             
                                & Vanilla AE & $236\times10^{-3}$              & $5.70\times10^{-1}$             & $9.63\times10^{-1}$      
                                             & $18.9\times10^{-8}$             & $10.7\times10^{-1}$             & $8.01\times10^{-1}$     \\
                                             
                                & Topo AE    & $232\times10^{-3}$              & $6.26\times10^{-1}$             & $9.66\times10^{-1}$     
                                             & $17.9\times10^{-8}$             & $1.89\times10^{-1}$             & $8.63\times10^{-1}$      \\
                                             
                                & t-SNE      & $9.88\times10^{-3}$             & $8.72\times10^{-1}$             & $9.99\times10^{-1}$     
                                             & $4.49\times10^{-8}$             & $510\times10^{-1}$              & $9.15\times10^{-1}$      \\
                                             
                                & UMAP       & $113\times10^{-3}$              & $8.38\times10^{-1}$             & $9.99\times10^{-1}$      
                                             & $5.66\times10^{-8}$             & $73.6\times10^{-1}$             & $9.05\times10^{-1}$      \\ \midrule

    \multirow{5}{*}{Car}        & Ours       & $11.4\times10^{-3}$             & $4.55\times10^{-1}$             & $9.05\times10^{-1}$      
                                             & $89.6\times10^{-9}$             & $4.31\times10^{-1}$             & $8.62\times10^{-1}$      \\
                                             
                                & Vanilla AE & $6.18\times10^{-3}$             & $5.35\times10^{-1}$             & $9.64\times10^{-1}$      
                                             & $89.5\times10^{-9}$             & $13.4\times10^{-1}$             & $9.06\times10^{-1}$     \\
                                             
                                & Topo AE    & $3.22\times10^{-3}$             & $5.50\times10^{-1}$             & $9.65\times10^{-1}$     
                                             & $8.06\times10^{-9}$             & $2.27\times10^{-1}$             & $9.78\times10^{-1}$      \\
                                             
                                & t-SNE      & $12.5\times10^{-3}$             & $7.35\times10^{-1}$             & $9.93\times10^{-1}$     
                                             & $69.0\times10^{-9}$             & $515\times10^{-1}$              & $9.14\times10^{-1}$      \\
                                             
                                & UMAP       & $10.8\times10^{-3}$             & $7.13\times10^{-1}$             & $9.91\times10^{-1}$      
                                             & $76.7\times10^{-9}$             & $67.3\times10^{-1}$             & $9.17\times10^{-1}$      \\ \midrule                                                
    \multirow{5}{*}{Chair}      & Ours       & $1.82\times10^{-3}$             & $3.91\times10^{-1}$             & $7.71\times10^{-1}$      
                                             & $13.3\times10^{-8}$             & $5.05\times10^{-1}$             & $5.25\times10^{-1}$      \\
                                             
                                & Vanilla AE & $2.13\times10^{-3}$             & $4.84\times10^{-1}$             & $9.32\times10^{-1}$      
                                             & $15.0\times10^{-8}$             & $9.66\times10^{-1}$             & $7.68\times10^{-1}$     \\
                                             
                                & Topo AE    & $1.83\times10^{-3}$             & $5.21\times10^{-1}$             & $9.38\times10^{-1}$     
                                             & $10.9\times10^{-8}$             & $1.89\times10^{-1}$             & $8.74\times10^{-1}$      \\
                                             
                                & t-SNE      & $1.16\times10^{-3}$             & $7.48\times10^{-1}$             & $9.94\times10^{-1}$     
                                             & $7.09\times10^{-8}$             & $507\times10^{-1}$              & $9.01\times10^{-1}$      \\
                                             
                                & UMAP       & $1.99\times10^{-3}$             & $6.98\times10^{-1}$             & $9.90\times10^{-1}$      
                                             & $11.0\times10^{-8}$             & $88.7\times10^{-1}$             & $8.64\times10^{-1}$      \\ \midrule                                       
    \multirow{5}{*}{Earphone}   & Ours       & $4.06\times10^{-3}$             & $5.04\times10^{-1}$             & $9.36\times10^{-1}$      
                                             & $9.57\times10^{-9}$             & $2.27\times10^{-1}$             & $9.41\times10^{-1}$      \\
                                             
                                & Vanilla AE & $21.8\times10^{-3}$             & $3.68\times10^{-1}$             & $8.71\times10^{-1}$      
                                             & $72.2\times10^{-9}$             & $9.14\times10^{-1}$             & $7.71\times10^{-1}$     \\
                                             
                                & Topo AE    & $23.9\times10^{-3}$             & $3.80\times10^{-1}$             & $8.75\times10^{-1}$     
                                             & $71.9\times10^{-9}$             & $1.74\times10^{-1}$             & $7.82\times10^{-1}$      \\
                                             
                                & t-SNE      & $17.6\times10^{-3}$             & $7.61\times10^{-1}$             & $9.94\times10^{-1}$     
                                             & $298\times10^{-9}$              & $513\times10^{-1}$              & $7.63\times10^{-1}$      \\
                                             
                                & UMAP       & $8.90\times10^{-3}$             & $7.33\times10^{-1}$             & $9.92\times10^{-1}$      
                                             & $305\times10^{-9}$              & $108\times10^{-1}$              & $6.83\times10^{-1}$      \\ \midrule

    \multirow{5}{*}{Lamp}       & Ours       & $7.41\times10^{-3}$             & $5.46\times10^{-1}$             & $8.91\times10^{-1}$      
                                             & $7.76\times10^{-8}$             & $2.39\times10^{-1}$             & $8.67\times10^{-1}$      \\
    
                                & Vanilla AE & $68.2\times10^{-3}$             & $4.91\times10^{-1}$             & $8.99\times10^{-1}$      
                                             & $106\times10^{-8}$              & $6.18\times10^{-1}$             & $5.80\times10^{-1}$      \\
                                
                                & Topo AE    & $70.3\times10^{-3}$             & $5.17\times10^{-1}$             & $9.10\times10^{-1}$      
                                             & $114\times10^{-8}$              & $2.62\times10^{-1}$             & $6.31\times10^{-1}$      \\

                                & t-SNE      & $55.6\times10^{-3}$             & $8.44\times10^{-1}$             & $9.97\times10^{-1}$      
                                             & $39.8\times10^{-8}$             & $509\times10^{-1}$              & $6.69\times10^{-1}$      \\
                                
                                & UMAP       & $54.0\times10^{-3}$             & $8.09\times10^{-1}$             & $9.96\times10^{-1}$      
                                             & $23.9\times10^{-8}$             & $92.1\times10^{-1}$             & $6.12\times10^{-1}$      \\ \midrule

    \multirow{5}{*}{Motorbike}  & Ours       & $15.9\times10^{-3}$             & $3.86\times10^{-1}$             & $8.81\times10^{-1}$      
                                             & $91.6\times10^{-8}$             & $3.57\times10^{-1}$             & $8.31\times10^{-1}$      \\
    
                                & Vanilla AE & $4.44\times10^{-3}$             & $3.80\times10^{-1}$             & $8.97\times10^{-1}$      
                                             & $10.5\times10^{-8}$             & $8.81\times10^{-1}$             & $9.20\times10^{-1}$      \\
                                
                                & Topo AE    & $4.77\times10^{-3}$             & $3.85\times10^{-1}$             & $8.96\times10^{-1}$      
                                             & $9.57\times10^{-8}$             & $2.06\times10^{-1}$             & $9.27\times10^{-1}$      \\

                                & t-SNE      & $9.63\times10^{-3}$             & $7.29\times10^{-1}$             & $9.94\times10^{-1}$      
                                             & $24.5\times10^{-8}$             & $549\times10^{-1}$              & $9.17\times10^{-1}$      \\
                                
                                & UMAP       & $7.88\times10^{-3}$             & $6.90\times10^{-1}$             & $9.92\times10^{-1}$      
                                             & $61.4\times10^{-8}$             & $94.3\times10^{-1}$             & $8.95\times10^{-1}$      \\ \midrule

    \multirow{5}{*}{Pistol}     & Ours       & $5.09\times10^{-3}$             & $4.76\times10^{-1}$             & $9.33\times10^{-1}$      
                                             & $3.67\times10^{-8}$             & $2.18\times10^{-1}$             & $9.31\times10^{-1}$      \\
    
                                & Vanilla AE & $7.40\times10^{-3}$             & $3.62\times10^{-1}$             & $8.48\times10^{-1}$      
                                             & $2.24\times10^{-8}$             & $6.30\times10^{-1}$             & $8.11\times10^{-1}$      \\
                                
                                & Topo AE    & $6.78\times10^{-3}$             & $3.90\times10^{-1}$             & $8.71\times10^{-1}$      
                                             & $1.80\times10^{-8}$             & $1.54\times10^{-1}$             & $8.29\times10^{-1}$      \\

                                & t-SNE      & $3.39\times10^{-3}$             & $8.03\times10^{-1}$             & $9.96\times10^{-1}$      
                                             & $20.0\times10^{-8}$             & $509\times10^{-1}$              & $7.78\times10^{-1}$      \\
                                
                                & UMAP       & $2.24\times10^{-3}$             & $7.73\times10^{-1}$             & $9.96\times10^{-1}$      
                                             & $16.3\times10^{-8}$             & $99.4\times10^{-1}$             & $8.31\times10^{-1}$      \\ \midrule

    \multirow{5}{*}{Rocket}     & Ours       & $24.8\times10^{-4}$             & $5.66\times10^{-1}$             & $9.67\times10^{-1}$      
                                             & $28.0\times10^{-9}$             & $3.09\times10^{-1}$             & $9.77\times10^{-1}$      \\
    
                                & Vanilla AE & $7.27\times10^{-4}$             & $5.94\times10^{-1}$             & $9.81\times10^{-1}$      
                                             & $8.54\times10^{-9}$             & $7.86\times10^{-1}$             & $9.91\times10^{-1}$      \\
                                
                                & Topo AE    & $4.08\times10^{-4}$             & $6.13\times10^{-1}$             & $9.81\times10^{-1}$      
                                             & $3.08\times10^{-9}$             & $1.89\times10^{-1}$             & $9.95\times10^{-1}$      \\

                                & t-SNE      & $66.8\times10^{-4}$             & $7.50\times10^{-1}$             & $9.95\times10^{-1}$      
                                             & $42.6\times10^{-9}$             & $625\times10^{-1}$              & $9.52\times10^{-1}$      \\
                                
                                & UMAP       & $52.9\times10^{-4}$             & $7.01\times10^{-1}$             & $9.93\times10^{-1}$      
                                             & $111\times10^{-9}$              & $112\times10^{-1}$              & $9.38\times10^{-1}$      \\ \midrule

    \multirow{5}{*}{Skateboard} & Ours       & $2.16\times10^{-3}$             & $7.03\times10^{-1}$             & $9.85\times10^{-1}$      
                                             & $29.8\times10^{-9}$             & $4.26\times10^{-1}$             & $9.80\times10^{-1}$      \\
    
                                & Vanilla AE & $1.92\times10^{-3}$             & $7.28\times10^{-1}$             & $9.94\times10^{-1}$      
                                             & $42.0\times10^{-9}$             & $12.4\times10^{-1}$             & $9.73\times10^{-1}$      \\
                                
                                & Topo AE    & $1.10\times10^{-3}$             & $8.07\times10^{-1}$             & $9.95\times10^{-1}$      
                                             & $3.39\times10^{-9}$             & $2.67\times10^{-1}$             & $9.96\times10^{-1}$      \\

                                & t-SNE      & $4.76\times10^{-3}$             & $8.08\times10^{-1}$             & $9.97\times10^{-1}$      
                                             & $20.7\times10^{-9}$             & $583\times10^{-1}$              & $9.73\times10^{-1}$      \\
                                
                                & UMAP       & $7.75\times10^{-3}$             & $7.62\times10^{-1}$             & $9.96\times10^{-1}$      
                                             & $160\times10^{-9}$              & $110\times10^{-1}$              & $9.16\times10^{-1}$      \\ \midrule

    \multirow{5}{*}{Table}      & Ours       & $7.15\times10^{-3}$             & $6.28\times10^{-1}$             & $9.74\times10^{-1}$      
                                             & $5.48\times10^{-8}$             & $3.45\times10^{-1}$             & $9.61\times10^{-1}$      \\
    
                                & Vanilla AE & $37.4\times10^{-3}$             & $5.46\times10^{-1}$             & $9.14\times10^{-1}$      
                                             & $12.0\times10^{-8}$             & $13.4\times10^{-1}$             & $8.05\times10^{-1}$      \\
                                
                                & Topo AE    & $40.1\times10^{-3}$             & $6.10\times10^{-1}$             & $9.20\times10^{-1}$      
                                             & $10.1\times10^{-8}$             & $2.35\times10^{-1}$             & $8.18\times10^{-1}$      \\

                                & t-SNE      & $12.1\times10^{-3}$             & $8.08\times10^{-1}$             & $9.98\times10^{-1}$      
                                             & $13.0\times10^{-8}$             & $493\times10^{-1}$              & $8.81\times10^{-1}$      \\
                                
                                & UMAP       & $11.7\times10^{-3}$             & $7.80\times10^{-1}$             & $9.96\times10^{-1}$      
                                             & $11.8\times10^{-8}$             & $107\times10^{-1}$              & $9.25\times10^{-1}$      \\ \bottomrule                                   
    \end{tabular}
}
\end{table}

\newpage

\hypertarget{Figure S2}{}
\begin{figure}[h]
    \centering
    \includegraphics[width=1\linewidth]{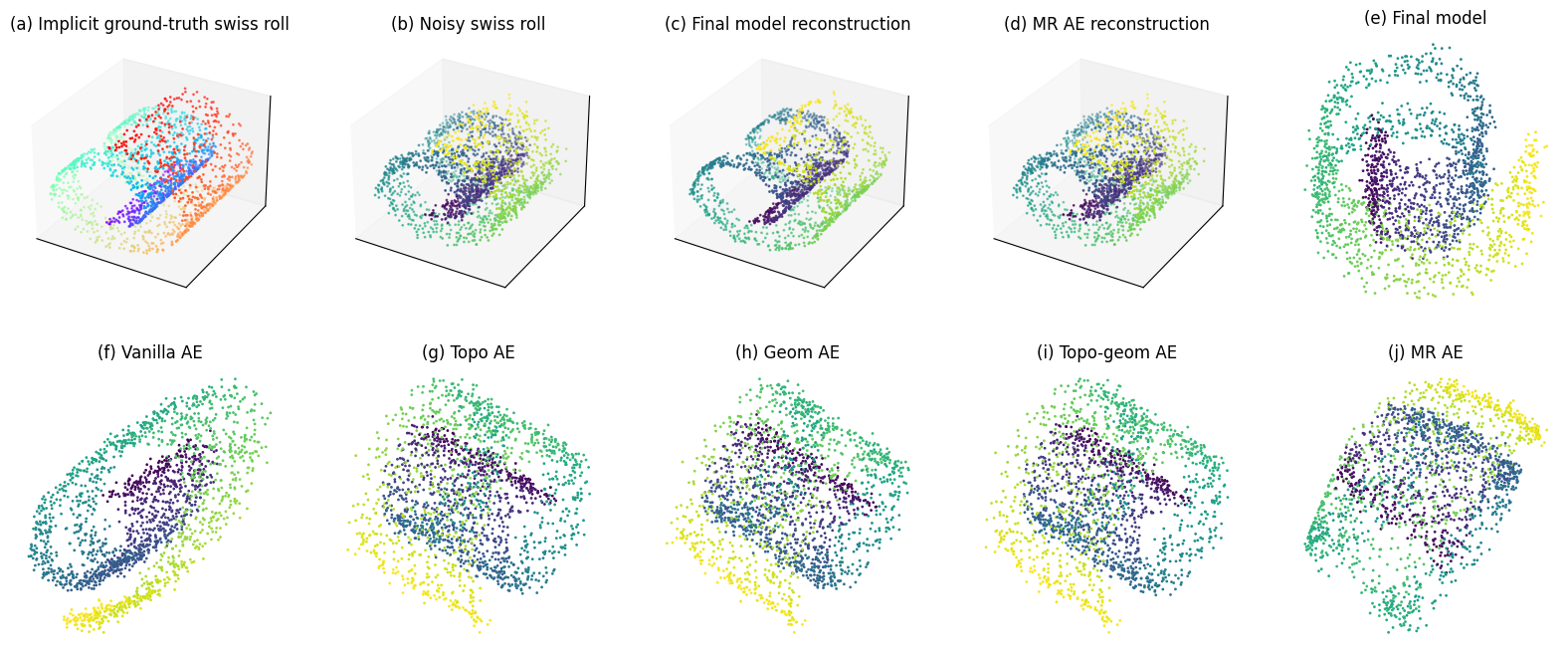}
    \caption{
    The visualization of our ablation study.
    (a) is the implicit ground-truth Swiss roll point cloud;
    (b) is the noisy 3D point cloud, which is the input data of our models;
    (c) and (d) are reconstructed latent manifolds with our Final model and MR AE, respectively;
    (e)-(j) are the embeddings of various models to compare in the ablation study.
    }
    \label{fig: ablation}
\end{figure}

\hypertarget{Table S2}{}
\begin{table}[h]
\caption
{
The evaluation metrics of our ablation study.
Evaluation and comparison are conducted in three groups:
(1) Noisy point clouds versus low-d embeddings,
(2) reconstructed manifolds versus low-d embeddings,
and (3) noisy point clouds versus reconstructed manifolds.
\textbf{Bold} values indicate winners from each group,
and \underline{underlined} are winners from the comparison between Topo-geom AE in "Point Cloud vs Embedding"
and the Final model in "Manifold vs Embedding".
}
\label{tab: ablation}
\resizebox{\textwidth}{!}
{
\begin{tabular}{cccccccc}
\toprule
\multirow{2}{*}{Comparison} & \multirow{2}{*}{Method} & \multicolumn{3}{c}{Local} & \multicolumn{3}{c}{Global} \\ \cline{3-8} 
                                                       &            & $\mathrm{KL_{0.1}}(\downarrow)$   & kNN$(\uparrow)$ & Trust$(\uparrow)$ & $\mathrm{KL_{100}}(\downarrow)$ & RMSE$(\downarrow)$ & Spear$(\uparrow)$ \\ \midrule

\multirow{6}{*}{P.C. vs E.}
                                                       & Final model & $\mathbf{1.95\times10^{-2}}$ & $\mathbf{5.37\times10^{-1}}$ &$\mathbf{9.03\times10^{-1}}$ 
                                                                     & $9.69\times10^{-8}$          & $3.56\times10^{-1}$          & $7.75\times10^{-1}$      \\

                                                       & Vanilla AE  & $2.54\times10^{-2}$          & $5.03\times10^{-1}$          & $9.02\times10^{-1}$      
                                                                     & $13.3\times10^{-8}$          & $11.9\times10^{-1}$          & $6.98\times10^{-1}$      \\
                                                       
                                                       & Topo AE     & $2.67\times10^{-2}$          & $5.25\times10^{-1}$          & $8.88\times10^{-1}$      
                                                                     & $9.80\times10^{-8}$          & $4.05\times10^{-1}$          & $7.84\times10^{-1}$      \\
                                                       
                                                       & Geom AE      & $2.44\times10^{-2}$          & $5.08\times10^{-1}$          & $8.93\times10^{-1}$      
                                                                     & $\mathbf{8.39\times10^{-8}}$ & $4.05\times10^{-1}$          & $\mathbf{7.93\times10^{-1}}$      \\
                                                       
                                                       & Topo-geom AE & $2.67\times10^{-2}$            & $5.30\times10^{-1}$          & $8.85\times10^{-1}$      
                                                                     & $\underline{9.62\times10^{-8}}$ & $4.06\times10^{-1}$          
                                                                     & $\underline{7.83\times10^{-1}}$      \\

                                                       & MR AE       & $2.78\times10^{-2}$          & $4.61\times10^{-1}$          & $8.82\times10^{-1}$     
                                                                     & $13.9\times10^{-8}$          & $\mathbf{1.72\times10^{-1}}$ & $7.13\times10^{-1}$      \\ \hline

\multirow{2}{*}{M. vs E.}               
                                                       & Final model & $\mathbf{\underline{2.00\times10^{-2}}}$ & $\mathbf{\underline{5.43\times10^{-1}}}$ 
                                                                     & $\mathbf{\underline{9.02\times10^{-1}}}$
                                                                     & $\mathbf{1.02\times10^{-7}}$ & $\mathbf{\underline{3.43\times10^{-1}}}$ 
                                                                     & $\mathbf{7.74\times10^{-1}}$ \\

                                                       & MR AE  & $2.54\times10^{-2}$          & $5.03\times10^{-1}$          & $9.02\times10^{-1}$
                                                                     & $1.33\times10^{-7}$          & $11.9\times10^{-1}$          & $6.98\times10^{-1}$      \\ \hline

\multirow{2}{*}{P.C. vs M.}                  
                                                       & Final model & $\mathbf{1.39\times10^{-3}}$ & $\mathbf{8.92\times10^{-1}}$ & $\mathbf{10.0\times10^{-1}}$ 
                                                                     & $\mathbf{8.56\times10^{-9}}$ & $\mathbf{2.54\times10^{-2}}$ & $\mathbf{9.93\times10^{-1}}$ \\

                                                       & MR AE  & $27.8\times10^{-3}$  & $4.61\times10^{-1}$ & $8.82\times10^{-1}$
                                                                     & $139\times10^{-9}$  & $17.2\times10^{-2}$ & $7.13\times10^{-1}$      \\ \bottomrule
\end{tabular}
}
\end{table}

\textbf{Comments.}
In our ablation study:
\begin{enumerate}
    \item "Final model" = MRL + AE + TopoReg + GeomReg;
    \item "Topo AE" = AE + TopoReg;
    \item "Geom AE" = AE + GeomReg;
    \item "Topo-geom AE" = AE + TopoReg + GeomReg;
    \item "MR AE" = MRL + AE.
\end{enumerate}

About the "mutual promotion effect":
when we introduce the Manifold Reconstruction Layer, the dimensionality reduction component (AutoEncoder + two regularizers) performs better, as depicted in a comparison between "Manifold vs Embedding - Final model" and "Point Cloud vs Embedding - Topo-geom AE" (there is no reconstructed manifold in the Topo-geom AE model). When we compare our final model with the MRL+AE model (the final model has two extra training regularizers) in the "Manifold vs Embedding" section, the final model also wins, suggesting that an enhanced dimensionality reduction can still improve the manifold reconstruction process during training.

\newpage

\section{Code and data}

The code and data can be found at \url{https://github.com/Thanatorika/mrtg}.

\section{Mathematical formulations}
\label{appendix: C}
\setcounter{theorem}{0}
\setcounter{definition}{0}
\renewcommand{\thetheorem}{C\arabic{theorem}}
\renewcommand{\thedefinition}{C\arabic{definition}}

\subsection{Assumptions of datasets}

\begin{definition} (\hypertarget{norm}{Norm})

    If $V$ is a real vector space, a \emph{norm on $V$} is a function from $V$ to $\mathbb{R}$, written $v\mapsto |v|$,
    satisfying the following properties:
    \begin{enumerate}
        \item Positivity: $|v|\le0$ for all $v\in V$, with equality if and  only if $v=0$.
        \item Homogeneity: $|cv|=|c||v|$ for all $c\in\mathbb{R}$ and $v\in V$.
        \item Triangular inequality: $|v+w|\le|v|+|w|$ for all $v,w\in V$.
    \end{enumerate}
    A vector space together with a specific choice of norm is called a \emph{normed linear space}.
\end{definition}

Example: $\mathbb{R}^n$ endowed with the \emph{Euclidean norm} defined by $|x|=\sqrt{x\cdot x}$ is a normed linear space.

\begin{definition} (\hypertarget{support}{Support of a function})

    If $f$ is any function on a topological space, the \emph{support of $f$}, denoted by $\mathrm{Supp}\ f$,
    is the closure of the set of points where $f$ is nonzero:
    $$
    \mathrm{Supp}\ f=\overline{\{p\in \mathcal{M}:f(p)\ne0\}}.
    $$
\end{definition}

\begin{definition} (\hypertarget{compact}{Compactness})

    A topological space $X$ is called \emph{compact} is every open cover of $X$ has a finite subcover.
    That is, $X$ is compact if for every collection $C$ of open subsets of $X$ such that
    $$X=\bigcup_{S\in C}S,$$
    there is a finite subcollection $F\subseteq C$ such that
    $$X=\bigcup_{S\in F}S.$$
\end{definition}

The Heine–Borel theorem:
For any subset $A$ of Euclidean space $\mathbb{R}^n$, $A$ is compact if and only if it is closed and bounded.

\begin{definition} (\hypertarget{Hausdorff measure}{Hausdorff measure})

    Suppose $(X,\rho)$ is a metric space. Let $S\subset X$, then the \emph{diameter} of $S$ is defined by
    $$
    \mathrm{diam}(S)=\max\{\rho(x,y):x,y\in S\}.
    $$
    Suppose $\delta>0$ is a real number,
    we define the \emph{Hausdorff measure} of dimension $d$ bounded by $\delta$ by
    $$
    H^d_\delta(S)=\inf\{\sum^\infty_{i=1}[\mathrm{diam}(U_i)]^d:\bigcup^\infty_{i=1}U_i\supseteq S, \mathrm{diam}(U_i)<\delta\},
    $$
    where the infimum is taken over all countable covers of $S$ by sets $U_i\subseteq X$ satisfying $\mathrm{diam}(U_i)<\delta$.
\end{definition}

The Hausdorff measure can be regarded as a generalized form of "volume".

\subsection{Manifold Reconstruction}

\begin{theorem} (Theoretical validation of the manifold fitting algorithm)

    Suppose $\mathcal{M}$ is the latent manifold, and $\sigma$ is the standard deviation of the Gaussian noise of our dataset.
    
    Assume the sample size $N=C_1\sigma^{-(d+3)}$, for a point $x$ such that $d(x,\mathcal{M})=O(\sigma)$,
    $F_{\mathcal{M}}$ (as defined in Equation \ref{FM}) provides an estimation of $x^*=\mathrm{argmin}_{x_{\mathcal{M}}\in\mathcal{M}}||x-x_{\mathcal{M}}||$,
    whose error can be bounded by
    $$
    ||F_{\mathcal{M}}(x)-x^*||_2\le C_2\sigma^2\mathrm{log}(1/\sigma)
    $$
    with probability at least $1-C_3\exp(-C_4\sigma^{-c})$, for some constant $c$, $C_1$, $C_2$, $C_3$, and $C_4$.
\end{theorem}

The manifold fitting algorithm is based on a statistical estimation of the distribution of noise data points. We assume the noisy input data to be a uniform distribution on the latent manifold + Gaussian noise, so when we obtain a noisy data point, we can get rid of noise interruption and estimate its corresponding manifold point by taking a weighted mean of its neighboring points. And the manifold fitting algorithm is mainly about how to compute these weights.

\subsection{Embedding the latent manifold}

\begin{definition} (\hypertarget{Immersion}{Immersion})

    Suppose $\mathcal{M}$ and $\mathcal{N}$ are manifolds, a smooth map $F:\mathcal{M}\to\mathcal{N}$ is called a \emph{smooth immersion} if its differential is injective at each point.
\end{definition}

\begin{definition} (\hypertarget{Embedding}{Embedding})

    A \emph{smooth embedding of $\mathcal{M}$ into $\mathcal{N}$} is a smooth immersion $F:\mathcal{M}\to\mathcal{N}$ that is also a topological embedding, i.e. a homeomorphism onto its image $F(\mathcal{M})\subseteq\mathcal{N}$ in the subspace topology.
\end{definition}

\begin{theorem} (\hypertarget{Whitney}{Strong Whitney Embedding Theorem})

    If $n>0$, every smooth $n$-manifold admits a smooth embedding into $\mathbb{R}^{2n}$.
\end{theorem}

According to strong Whitney embedding theorem,
in a representational space with $d \ge 2L$ dimensions,
there is guaranteed to be an topological embedding map;
while in representational spaces with $d<2L$,
we can still promote the similarity of topological and geometric properties between the latent manifold and the representation.

\subsection{Diffeomorphism and Persistent Homology}

\begin{definition} (\hypertarget{Homology}{Homology})

    A \emph{chain complex} is a sequence $(C_\bullet,d_\bullet)$ of abelian groups $C_n$ and group homomorphisms $d_n$ such that the composition of any two consecutive maps is zero:
    $$
    C_\bullet:\cdots\xrightarrow{} C_{n+1}\xrightarrow{d_{n+1}}C_n\xrightarrow{d_n}C_{n-1}\xrightarrow{d_{n-1}}\cdots, d_n\circ d_{n+1}=0.
    $$
    The $n$-th group of \emph{cycles} is defined as the kernel subgroup $Z_n=\mathrm{ker}\ d_n=\{c\in C_n\ |\ d_n(c)=0\}$,
    and the $n$-th group of \emph{boundaries} is given by the image subgroup $B_n=\mathrm{im}\ d_{n+1}=\{d_{n+1}(c)\ |\ c\in C_{n+1}\}$.
    The $n$-th \emph{homology group} $H_n$ of this chain complex is defined as the quotient group $H_n=Z_n/B_n$.
\end{definition}

Roughly speaking, homology is some "feature" of a topological object, and different topological objects (e.g. manifolds) can be distinguished apart by their different homology. Homology can be computed in many forms, and Persistent Homology is a prevalent method for computing the homology features of point clouds.

\begin{definition} (\hypertarget{Homeomorphism}{Homeomorphism})

    A continuous bijective map $F:X\to Y$ with continuous inverse is called a \emph{homeomorphism}.
    If there exists a homeomorphism from $X$ to $Y$, we say that $X$ and $Y$ are \emph{homeomorphic}.
\end{definition}

From a topological viewpoint, two homeomorphic topological spaces are the same.
An illustration is the famous joke of “mathematicians cannot tell the difference between a doughnut and a coffee mug”.

\begin{theorem} 
    Homeomorphic manifolds have the same homology groups.
\end{theorem}

\begin{definition} (\hypertarget{Diffeomorphism}{Diffeomorphism})

    If $\mathcal{M}$ and $\mathcal{N}$ are smooth manifolds, a \emph{diffeomorphism from $\mathcal{M}$ to $\mathcal{N}$} is a smooth bijective map $F:\mathcal{M}\to\mathcal{N}$ that has a smooth inverse.
    We say that \emph{$\mathcal{M}$ and $\mathcal{N}$ are diffeomorphic} if there exists a diffeomporphism between them. 
\end{definition}

\begin{definition} (Simplicial homology)

    A \emph{simplicial complex} $\mathfrak{K}$ is a set composed of points, line segments, triangles, and their $n$-dimensional counterparts, i.e. a high-dimensional analogy of a graph.

    Let $C_d(\mathfrak{K})$ denote the vector spacr generated over $\mathbb{Z}_2$ whose elements are the $d$-simplices in $\mathfrak{K}$.
    For $\sigma=(v_0,\cdots,v_d)\in\mathfrak{K}$, let $\partial_d:C_d(\mathfrak{K})\to C_{d-1}(\mathfrak{K})$ be the boundary homomorphism defined by
    $$
    \partial_d(\sigma)=\sum^d_{i=0}(-1)^i(v_0,\cdots,v_{i-1},v_{i+1},\cdots,v_d).
    $$
    The $d$-th homology group $H_d(\mathfrak{K})$ is defined as the quotient group $H_d(\mathfrak{K})=\mathrm{ker}\ \partial_d/\mathrm{im}\ \partial_{d+1}$.
\end{definition}

\begin{definition} (\hypertarget{Persistent Homology}{Persistent homology})

    Let $\emptyset=\mathfrak{K_0}\subseteq\mathfrak{K}_1\subseteq\cdots\subseteq\mathfrak{K}_{m-1}\subseteq\mathfrak{K}_m=\mathfrak{K}$ be a nested sequence of simplicial complexes, called \emph{filtration}.
    
    The Vietoris-Rips filtration we discussed in the paper induces a homomorphism between corresponding homology groups: $f^{i,j}_d:H_d(\mathfrak{K}_i)\to H_d(\mathfrak{K}_j)$.
    This homomorphism yields a sequence of homology groups
    $$
    0=H_d(\mathfrak{K}_0)\xrightarrow{f^{0,1}_d}H_d(\mathfrak{K}_1)\xrightarrow{}\cdots\xrightarrow{f^{m-2,m-1}_d}H_d(\mathfrak{K}_{m-1})\xrightarrow{f^{m-1,m}_d}H_d(\mathfrak{K}_{m})=H_d(\mathfrak{K})
    $$
    for every dimension $d$. Given indices $i\le j$, the $d$-th \emph{persistent homology group} is defined as
    $$
    H^{i,j}_d=\mathrm{ker}\ \partial_d(\mathfrak{K}_i)/(\mathrm{\mathrm{im}\ \partial_{d+1}(\mathfrak{K}_j)}\cap\mathrm{ker}\ \partial_d(\mathfrak{K}_i)).
    $$
\end{definition}

\begin{definition} (\hypertarget{Persistent diagram}{Persistent diagrams})

    A filtration is often associated with weights $w_0\le w_1\le\cdots\le w_{m-1}\le w_m$, e.g. the pairwise distances in a point cloud.
    
    We can use these values to compute topological feature representations known as \emph{persistent diagrams}:
    for each dimension $d$ and each pair $i\le j$, one stores a pair $(a,b)=(w_i,w_j)\in\mathbb{R}^2$ with multiplicity
    $$
    \mu^{(d)}_{i,j}=(\beta_d^{i,j-1}-\beta^{i,j}_d)-(\beta^{i-1,j-1}_d-\beta^{i-1,j}_d),
    $$
    where $\beta^{i,j}_d=\mathrm{rank}\ H^{i,j}_d$ is the $d$-th \emph{persistent Betti number}.
    The resulting set of points is called the $d$-th \emph{persistent diagram} $\mathcal{D}_d$.
\end{definition}

Intuitively, Persistent Homology is computed by treating point clouds as simplices (high-dimensional analogy of graphs), and extracting the most enduring or "persistent" connectivity features within a increasing distance scale.

\subsection{Isometry and Relaxed Distortion Measure}

\begin{definition} (\hypertarget{Riemannian metric}{Riemannian manifold})

    A \emph{Riemannian metric on $\mathcal{M}$} is a smooth symmetric covariant 2-tensor field on $\mathcal{M}$ that is positive definite at each point.
    A \emph{Riemannian manifold} is a pair $(\mathcal{M},g)$, where $\mathcal{M}$ is a smooth manifold and $g$ is a Riemannian metric on $\mathcal{M}$.

    In any smooth local coordinates $(x^i)$, a Riemannian metric can be written as
    $$
    g=g_{ij}dx^i\otimes dx^j,
    $$
    where $(g_{ij})$ is a symmetric positive definite matrix of smooth functions. 
\end{definition}

Roughly speaking,
a Riemannian manifold is a smooth manifold equipped with a metric system, 
such that we can measure the local geometric properties around each point,
and how they vary along different curves.

\begin{definition} (\hypertarget{pullback metric}{Pullback metric})

    Suppose $\mathcal{M},\mathcal{N}$ are smooth manifolds, $g$ is a Riemannian metric on $\mathcal{N}$, and $F:\mathcal{M}\to\mathcal{N}$ is smooth.
    The pullback $F^*g$ is a smooth 2-tensor field on $\mathcal{M}$. If it is positive definite, it is a Riemannian metric on $\mathcal{M}$, called the \emph{pullback metric} determined by $F$.
\end{definition}

Intuitively,
a pullback operation is a "translation" of the Riemannian metric system from $\mathcal{N}$ to $\mathcal{M}$ (in the backward direction of $F:\mathcal{M}\to\mathcal{N}$, thus the name),
where the quality of the translator $F$ determines what results $\mathcal{N}$ will get.

\begin{definition} (\hypertarget{Isometry}{Isometry})

    If $(\mathcal{M},g)$ and $(\mathcal{N},h)$ are both Riemannian manifolds, a smooth map $F:\mathcal{M}\to\mathcal{N}$ is called a \emph{(Riemannian) isometry} if it is a diffeomorphism that satisfies $F^*h=g$. $F$ is called a \emph{scaled isometry} if it is a diffeomorphism that satisfies $F^*h=c\cdot g$ for some constant $c$.
\end{definition}

If $F$ is an isometry,
then $\mathcal{M}$ gets an identical metric system $g=F^*h$;
if $F$ is a scaled isometry,
then $\mathcal{M}$ gets a metric system from a giant or dwarf society $g=C\cdot F^*h$,
where all the metrics are scaled to some constant number.

\section{Datasets, experiments, metrics}

\subsection{Datasets formation}
\label{Appendix: datasets}

\textbf{Noisy Swiss roll with a hole.}
The dataset is a noisy 3D point cloud with the shape of a rolled surface with a hole on it.
It is based on the classic "Swiss roll" dataset in manifold learning,
which is widely adopted to test if the method can recognize and preserve the global shape (topology) of the surface,
rather than just projecting stacked points onto 2-dimensional space.
The implicit clean point cloud is generated with the "make\_Swiss\_roll" function in the Python package Scikit-learn \cite{scikit-learn},
with "n\_samples=2000" and "hole=True" settings.
The point values are then scaled to $[0,1]$, and added with a 3-dimensional Gaussian random noise with $\sigma=0.02$.
Colors are generated with the data points, indicating the "angle" of the rolled surface.

\textbf{Noisy mammoth.}
The dataset is a noisy 3D point cloud with the shape of a prehistory animal: the mammoth.
It is based on the "mammoth" dataset first shown in the "Understanding UMAP" project \cite{Mammoth}.
The clean mammoth dataset is retrieved from \url{https://github.com/PAIR-code/understanding-umap/blob/master/raw_data/mammoth_3d.json}.
The point values are then scaled to $[0,1]$, and added with a 3-dimensional Gaussian random noise with $\sigma=0.02$.
Colors are the same values as the "height" of the mammoth object. 

\textbf{PartNet dataset.}
This dataset is composed of 12 3D point clouds indicating different real-life objects,
and is a subset of the PartNet \cite{PartNet} dataset in 3D vision studies.
The dataset is part of the "sem\_seg\_h5.zip" file retrieved from \url{https://huggingface.co/datasets/ShapeNet/PartNet-archive}.
We select 12 representative objects to form our testing set.
Colors indicate semantic segmentation labels of each object.

\textbf{Spheres dataset.}
This dataset is composed of 9 distinct high-dimensional spheres, with one large sphere surrounding 8 smaller spheres.
The spheres are points in 101-dimensional Euclidean space, and intrinsically of 100 dimensions.
The dataset is generated with code from \url{https://github.com/aidos-lab/pytorch-topological/tree/main}.
Colors indicate different entities of spheres.

\subsection{Experiment details}
\label{Appendix: training}

\textbf{3D point clouds.}
For 3D point clouds, the input dimension is 3, and the output dimension is 2.
We build our AutoEncoder with a "3-2-2" two-layer encoder and a "2-2-3" two-layer decoder,
where each layer is a linear layer with a GELU activation function.
The last layer of the decoder is followed by a sigmoid function. 

Data values are normalized to $[0,1]$ scale before training.
We initialize the parameters of the Manifold Reconstruction Layer to be $r_0=1.0, r_1=0.01, r_2=1.0$, 
the AutoEncoder loss is MSE loss,
and the Topological Regularizers are computed with 1-dimensional Vietoris-Rips Complexes.
Our models and corresponding AutoEncoders are trained for 100 epochs with batch size 128 and learning rate $\eta=1\times10^{-3}$.

For the Swiss roll dataset, our loss term weights are $\lambda_{AE} = 1, \lambda_{topo} = 1, \lambda_{geom} = 5$;
for the mammoth dataset, they are $\lambda_{AE} = 1, \lambda_{topo} = 5\times10^{-1}, \lambda_{geom} = 5\times10^{-1}$;
and for the PartNet dataset, we set them as $\lambda_{AE} = 1, \lambda_{topo} = 1\times10^{-2}, \lambda_{geom} = 1\times10^{-2}$.

\textbf{High-dimensional point clouds.}
For 100-D spheres,
the input dimension is 101, and the output dimension is 2.
We build our AutoEncoder with a "101-64-32-16-8-4-2" encoder, and a "2-4-8-16-32-64-101" decoder,
where each layer is a linear layer with a GELU activation function.
The last layer of the decoder is followed by a sigmoid function.

Data values are of scale $[0,1]$. 
We initialize the parameters of the Manifold Reconstruction Layer to be $r_0=1.0, r_1=0.01, r_2=1.0$, 
the AutoEncoder loss is MSE loss,
and the Topological Regularizers are computed with 1-dimensional Vietoris-Rips Complexes.
Our models and corresponding AutoEncoders are trained for 100 epochs with batch size 128 and learning rate $\eta=1\times10^{-3}$.
Our loss term weights are $\lambda_{AE} = 1, \lambda_{topo} = 1, \lambda_{geom} = 1$.

\subsection{Evaluation metrics}
\label{Appendix: metrics}

The evaluation metrics are adopted from \cite{TopoAE,GeomAE}.
There are three local metrics $\mathrm{KL_{0.1}}$, kNN, and Trust;
and three global metrics $\mathrm{KL_{100}}$, RMSE, and Spear.

\begin{enumerate}
    \item The $\mathrm{KL_\sigma}$ metrics measure the Kullback-Leibler divergence between probability distributions of the original data set and embedding, where the parameter $\sigma$ defines a length scale.
    \item The kNN metric measures the proportion of nearest neighbors in the embedding that are also nearest neighbors in the original data set \cite{kNN}.
    \item The Trust metric is short for trustworthiness, which measures to what extent the $k$-nearest neighbors of a point are preserved when going from the 
          original dataset to the embedding \cite{Trust}.
    \item The RMSE is the root mean square error between the distance matrix of the original data set and the embedding.
    \item The Spear metric measures the Spearman correlation between the distances of all pairs of original and embedding points \cite{Spear}.
\end{enumerate}

All the metric scores are averaged over 10 embeddings, which are repetitively trained with the same hyperparameter settings and input datasets.
For kNN and Trust metrics that require a parameter $k$ for the number of nearest neighbors to consider,
the values are further averaged over a range of $k$'s from 10 to 100 in steps of 10, as proposed in \cite{TopoAE}.

\section{Supplemental experiments}

\setcounter{figure}{0}
\renewcommand{\thefigure}{E\arabic{figure}}

\setcounter{table}{0}
\renewcommand{\thetable}{E\arabic{table}}

\subsection{Comparison with Isomap}

We mentioned Isomap \cite{Isomap} as an alternative manifold learning approach in our introduction,
but did not include it in our main experiments as an opponent method.
It was because the performance of Isomap is not as competitive as other components in general,
and the training speed is relatively low when scaling up our experiment data sizes.

Here we perform experiments with a default-setting Isomap method on
Swiss roll, Mammoth, Spheres, and the "table" object from the PartNet dataset,
and compare its embedding visualizations and metrics evaluation with our proposed method.
The steps for metrics computation are the same as in the main text.

Figure \hyperlink{Figure E1}{E1} and Table \hyperlink{Table E1}{E1} are the visualization results and evaluation metrics.
We can see that our low-dimensional embeddings better preserved local and global structures than Isomap,
whereas exceeding it in most of the evaluation metrics comparisons.

\begin{figure}[t]
    \centering
    \includegraphics[width=1.0\linewidth]{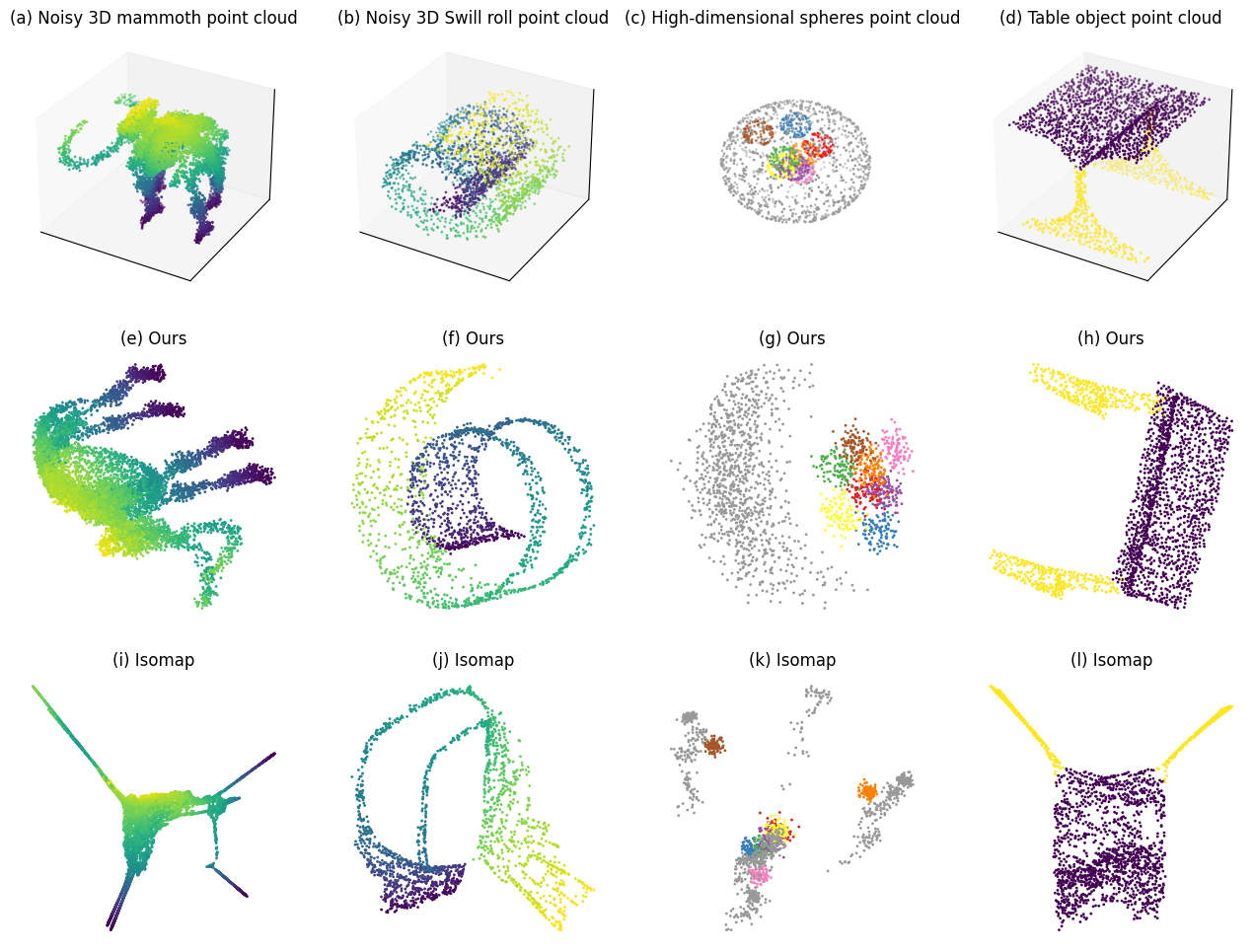}
    \caption{Visualization results with Isomap, compared to our method.}
\end{figure}
\hypertarget{Figure E1}{}

\hypertarget{Table E1}{}
\begin{table}[h]
\caption{
         Local and global metrics for our method and Isomap.
         \textbf{Bold} values are winners.
        }
\resizebox{\textwidth}{!}
{
    \begin{tabular}{@{}cccccccc@{}}
    \toprule
    \multirow{2}{*}{Dataset} & \multirow{2}{*}{Method} & \multicolumn{3}{c}{Local} & \multicolumn{3}{c}{Global} \\ \cmidrule(l){3-8}
                                &            & $\mathrm{KL_{0.1}}(\downarrow)$ & kNN$(\uparrow)$ & Trust$(\uparrow)$ & $\mathrm{KL_{100}}(\downarrow)$ & RMSE$(\downarrow)$ & Spear$(\uparrow)$ \\ \midrule

    \multirow{2}{*}{Mammoth}    & Ours       & $\mathbf{2.15\times10^{-3}}$    & $\mathbf{4.93\times10^{-1}}$    & $\mathbf{9.61\times10^{-1}}$      
                                             & $\mathbf{1.03\times10^{-8}}$    & $1.63\times10^{-1}$             & $\mathbf{9.76\times10^{-1}}$      \\
    
                                & Isomap     & $4.05\times10^{-3}$             & $3.48\times10^{-1}$             & $9.60\times10^{-1}$      
                                             & $2.27\times10^{-8}$             & $\mathbf{1.42\times10^{-1}}$    & $9.44\times10^{-1}$      \\ \midrule

    \multirow{2}{*}{Swiss roll} & Ours       & $\mathbf{2.19\times10^{-2}}$    & $4.69\times10^{-1}$             & $8.98\times10^{-1}$      
                                             & $\mathbf{1.12\times10^{-7}}$    & $\mathbf{2.55\times10^{-1}}$    & $\mathbf{7.60\times10^{-1}}$      \\
                                             
                                & Isomap     & $2.83\times10^{-2}$             & $\mathbf{6.60\times10^{-1}}$    & $\mathbf{9.80\times10^{-1}}$      
                                             & $1.67\times10^{-7}$             & $5.63\times10^{-1}$             & $4.74\times10^{-1}$      \\ \midrule

    \multirow{2}{*}{Spheres}    & Ours       & $4.78\times10^{-1}$             & $\mathbf{1.76\times10^{-1}}$    & $\mathbf{5.78\times10^{-1}}$      
                                             & $10.1\times10^{-7}$             & $\mathbf{2.81\times10^{1}}$              & $-1.85\times10^{-2}$      \\

                                & Isomap     & $\mathbf{4.36\times10^{-1}}$    & $1.67\times10^{-1}$             & $5.60\times10^{-1}$      
                                             & $\mathbf{6.73\times10^{-7}}$    & $27.3\times10^{1}$              & $\mathbf{2.44\times10^{-2}}$      \\ \midrule                           
                                             
    \multirow{2}{*}{Table}      & Ours       & $7.15\times10^{-3}$             & $\mathbf{6.28\times10^{-1}}$    & $\mathbf{9.74\times10^{-1}}$      
                                             & $5.48\times10^{-8}$             & $\mathbf{3.45\times10^{-1}}$    & $\mathbf{9.61\times10^{-1}}$      \\

                                & Isomap     & $\mathbf{5.84\times10^{-4}}$    & $5.39\times10^{-1}$             & $9.36\times10^{-1}$      
                                             & $\mathbf{3.90\times10^{-8}}$    & $3.57\times10^{-2}$             & $9.60\times10^{-1}$      \\ \bottomrule
    \end{tabular}
}
\end{table}

\subsection{Robustness under different data noise levels}

Here we perform an ablation experiment on different artificial noise levels of the Swiss roll dataset.
We define two categories of evaluations: the "loss" is the difference between the current data and the noisy input, and the "error" is the difference between the current data and the clean ground truth.

All the data curation, model implementation, and training hyperparameter settings are the same as in the main text, except that the artificial Gaussian introduced to the Swiss roll dataset now form a 11-entry staircase sequence with standard deviations ranging from $[0,0.05]$ with a step $0.05$.

\begin{figure}[h]
    \centering
    \includegraphics[width=1\linewidth]{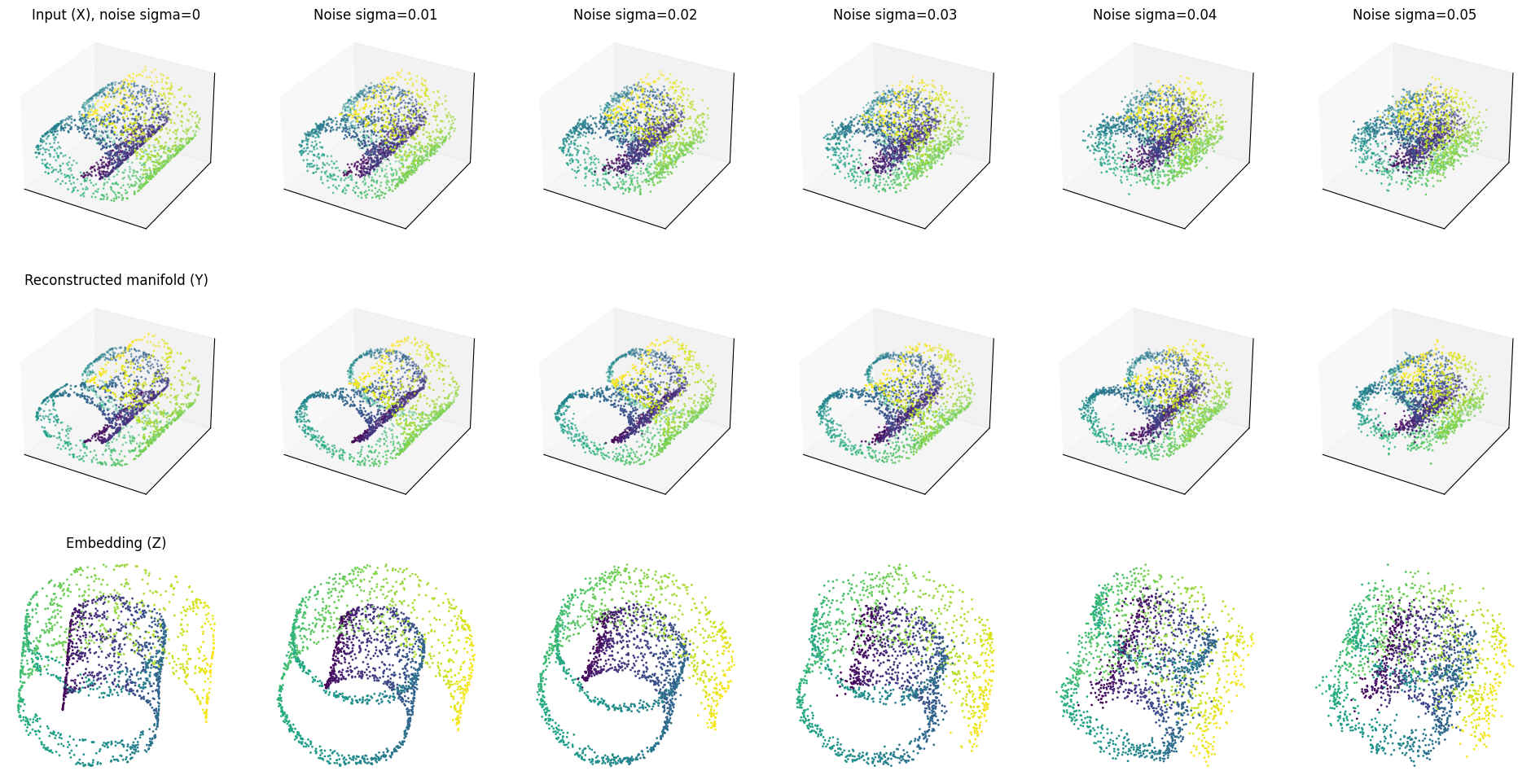}
    \caption{Visualization of different noise levels of the Swiss roll dataset}
\end{figure}
\hypertarget{Figure E2}{}

Figure \hyperlink{Figure E2}{E2} shows the input datasets, reconstructed manifolds, and 2-dimensional embeddings of different noise levels.
We can see that although the increasing noise levels confused the Swiss roll structures, our model still managed to reconstruct a fairly recognizable manifold structure, and embeds it with faithful local and global shapes.

\begin{figure}[h]
    \centering
    \includegraphics[width=1\linewidth]{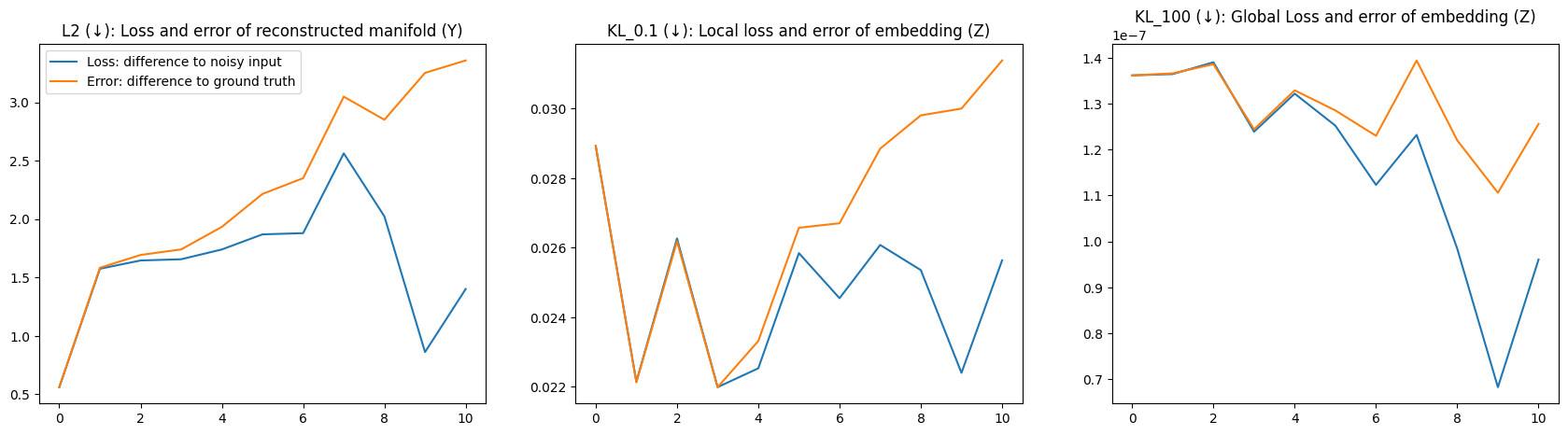}
    \caption{From left to right:
    The loss and error of the reconstructed manifold $Y$, with values as $L_2$ loss;
    the local loss and error of the embedding $Z$, with values as $\mathrm{KL}_{0.1}$;
    and the global loss and error of the embedding $Z$, with values as $\mathrm{KL}_{100}$.
    }
\end{figure}
\hypertarget{Figure E3}{}

Figure \hyperlink{Figure E3}{E3} shows the deviations between the ground truth Swiss roll, the noisy datasets, and the reconstructed manifolds and embeddings.

From the curves we can see that the differences between reconstructed manifolds and noisy inputs are relatively steady, while the differences between reconstructed manifolds and the ground truth continue to increase, both as expected. Similar trends hold for the comparisons between embeddings and noisy/ground truth Swiss rolls.

It is kind of an interesting discovery that the local and global metrics for the embeddings seem to improve with increasing noise levels.
We suspect the reason could be the increased noise levels confusing the specific manifold structures with a simpler Gaussian distribution, thus making the noisy input's geometric and topological structures more trivial, and easier to be passed down through the AutoEncoder model.

\subsection{Robustness under different data scales}

\subsubsection{Computational cost of the model}

In the three modifications of the AutoEncoder model, the Topological Regularizer is the main computational bottleneck.

The algorithms of the Manifold Reconstruction Layer \cite{YSLY23} and the Geometric Regularizer \cite{IsoAE} are both non-iterative, but only require matrix operations with the mini-batch tensor of size $[D,n]$, where $D$ is the dimensionality, and $n$ is the batch size.

The computation of the Topological Regularizer \cite{TopoAE} utilizes Persistent Homology, which includes the construction of Vietoris-Rips simplices. For an input sample with batch size $n$ and simplex dimensionality (a hyperparameter) $s$, the size of a simplex is $O(n^s)$, and there are $\tbinom{n}{s+1}$ possible $s$-simplices.

In summary, the computational cost of our model is dominant by the construction of Vietoris-Rips simplices, which increases exponentially with the predefined simplex dimensionality.
However in real life applications, we generally do not cope with topological data features of dimensionality greater than $3$ (e.g. a hole in a solid 4 dimensional object), so under most circumstances we can set our simplex dimensionality in $[0,1,2]$. If we only need to perform clustering on our data, then $0$-dimensional simplices should work.

There are several approaches to reduce the computational complexity of Persistent Homology, for example, Ripser \cite{Ripser} utilizes the sparsity of coboundary matrices.

\subsubsection{Scaling to larger sample sizes}

Here we perform an ablation experiment on different sample sizes of the Swiss roll dataset.We define two categories of evaluations: the "loss" is the difference between the current data and the noisy input, and the "error" is the difference between the current data and the clean ground truth.

All the data curation, model implementation, and training hyperparameter settings are the same as in the main text, except that the input sample sizes of the Swiss roll dataset now form a 10-entry staircase sequence with data sizes ranging from $[500,5000]$ with a step $500$.

\begin{figure}[h]
    \centering
    \includegraphics[width=1\linewidth]{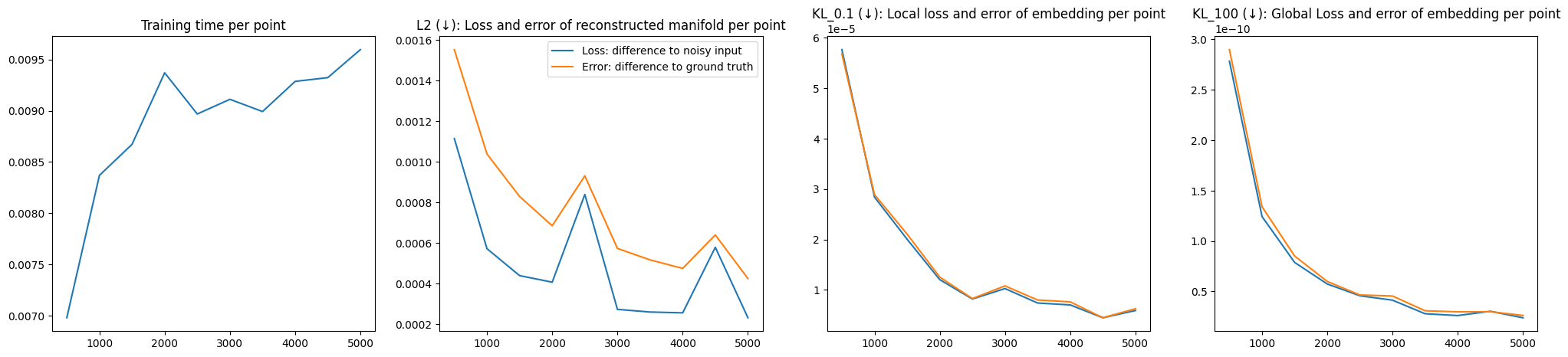}
    \caption{From left to right:
    The training time per input data point with increasing sample sizes;
    the loss and error of the reconstructed manifold $Y$, with values as $L_2$ loss;
    the local loss and error of the embedding $Z$, with values as $\mathrm{KL}_{0.1}$;
    and the global loss and error of the embedding $Z$, with values as $\mathrm{KL}_{100}$.}
\end{figure}
\hypertarget{Figure E4}{}

Figure \hyperlink{Figure E4}{E4} shows the training time durations, deviations between the ground truth Swiss roll, the noisy datasets, and the reconstructed manifolds and embeddings, each averaged by sample sizes.

Averaged training time for each data point seem to increase with increased data sizes, but with an acceptable growth rate. It's our pleasure to see the averaged losses and errors all go down with increased data sizes with a decent rate.

\subsubsection{Scaling to higher dimensional data}

Here we perform an ablation experiment on different input dimensions of the Spheres dataset.

All the data curation, model implementation, and training hyperparameter settings are the same as in the main text, except that the input sample sizes of the Swiss roll dataset now form a 6-entry staircase sequence with input dimensions $[4,8,16,32,64,128]$, so the spheres are of intrinsic dimensionality $[3,7,15,31,63,127]$.

\begin{figure}[h]
    \centering
    \includegraphics[width=1\linewidth]{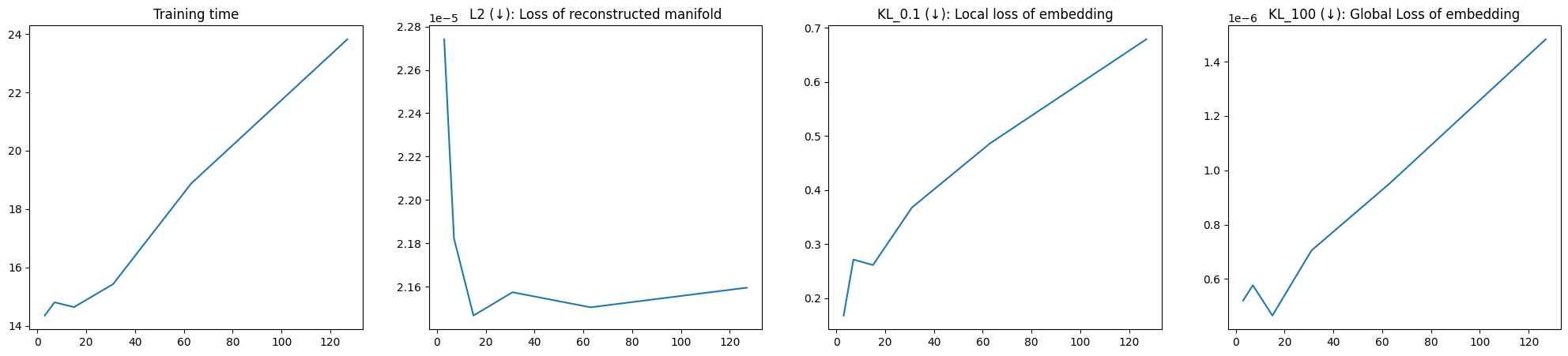}
    \caption{From left to right:
    The training time with increasing input dimensions;
    the loss of the reconstructed manifold $Y$, with values as $L_2$ loss;
    the local loss of the embedding $Z$, with values as $\mathrm{KL}_{0.1}$;
    and the global loss of the embedding $Z$, with values as $\mathrm{KL}_{100}$.}
\end{figure}
\hypertarget{Figure E5}{}

Figure \hyperlink{Figure E5}{E5} shows the training time durations, deviations between the input Spheres dataset and the reconstructed manifolds and embeddings.

Training times, local ($\mathrm{KL}_{0.1}$) losses, and global ($\mathrm{KL}_{100}$) losses all seem to increase linearly with increasing input dimensions, while the $L_2$ losses of the Manifold Reconstruction Layer first go down, then remain stable.

\subsection{Robustness under different hyperparameter settings}

Here we perform a grid search experiment on different hyperparameter values of $\lambda_{topo}$ and $\lambda_{geom}$, with Swiss roll and mammoth datasets.

All the data curation, model implementation, and training hyperparameter settings are the same as in the main text, except that the training loss hyperparameters $\lambda_{topo}$ and $\lambda_{geom}$ now form a 100-entry grid with each axis having values from $[0,0.01,0.05,0.1,0.5,1,5,10,50,100]$.

\begin{figure}[t]
    \centering
    \includegraphics[width=1\linewidth]{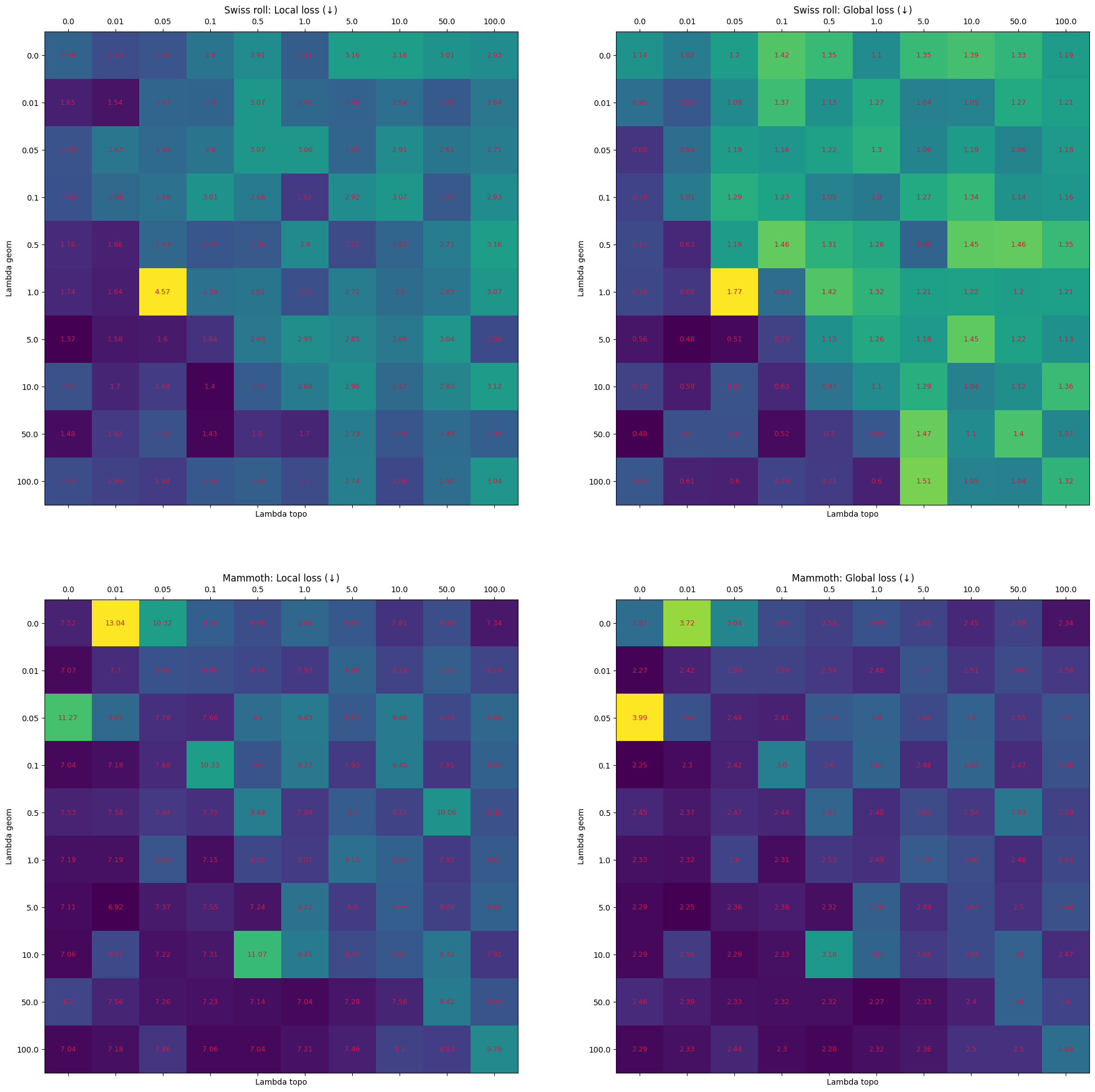}
    \caption{Top left: the local loss of the embeddings of Swiss rolls, with values as $\mathrm{KL}_{0.1}$.
    Top right: the global loss of the embeddings of Swiss rolls, with values as $\mathrm{KL}_{100}$.
    Bottom left: the local loss of the embeddings of mammoths, with values as $\mathrm{KL}_{0.1}$.
    Bottom right: the global loss of the embeddings of mammoths, with values as $\mathrm{KL}_{100}$.
    }
\end{figure}
\hypertarget{Figure E6}{}

Figure \hyperlink{Figure E6}{E6} shows the deviations between the input dataset and the embeddings.
We can see that overall the evaluation metrics are good in the left-bottom corner of the heatmap,
that is when we have a relatively small $\lambda_{topo}$ and a relatively large $\lambda_{geom}$.
This trend holds for both the Swiss roll and mammoth datasets.

We acknowledge that the optimization of hyperparameters in our model,
especially balancing the training loss coefficients $\lambda_{topo}$ and $\lambda_{geom}$ is an important and challenging problem, and requires further investigation in the future.
A potential direction may be to construct a more suitable metric to evaluate the performance of our dimensionality reduction module with noisy inputs,
and utilize hyperparameter tuning tools like Ray Tune \cite{Tune} to search for an optimal combination of hyperparameters.

\end{document}